\RequirePackage{fix-cm}
\documentclass[twocolumn]{svjour3}          
\smartqed  
\usepackage{graphicx}
\usepackage{amsmath,amssymb} 
\usepackage{color}
\usepackage{graphicx}
\usepackage{amsmath}
\usepackage{amssymb}
\usepackage{booktabs}
\usepackage{tabularx}
\usepackage{multirow}
\usepackage{xspace}

\DeclareGraphicsExtensions{.pdf}

\usepackage[pagebackref=true,breaklinks=true,letterpaper=true,colorlinks,bookmarks=false]{hyperref}

\usepackage{algorithm}
\usepackage[noend]{algpseudocode}
\makeatletter
\def\BState{\State\hskip-\ALG@thistlm}
\makeatother

\usepackage[dvipsnames]{xcolor}
\newcolumntype{Y}{>{\raggedright\arraybackslash}X} 

\usepackage{natbib}

\newcommand{\rtext}[2]{\parbox[t]{2mm}{\multirow{#2}{*}{\rotatebox[origin=c]{90}{#1}}}}

\makeatletter
\DeclareRobustCommand\onedot{\futurelet\@let@token\@onedot}
\def\@onedot{\ifx\@let@token.\else.\null\fi\xspace}

\def\eg{\emph{e.g}\onedot} 
\def\ie{\emph{i.e}\onedot} 
\def\cf{\emph{c.f}\onedot}

\makeatother

\usepackage{pifont}
\newcommand{\cmark}{\ding{51}}%
\begin{document}

\title{End-to-end Learning of Deep Visual Representations\\ for Image Retrieval}

\titlerunning{End-to-end Learning Deep Visual Representations for Image Search}

\authorrunning{A. Gordo \and J. Almaz\'an \and J. Revaud \and D. Larlus}

\author{Albert Gordo \and Jon Almaz\'an \and Jerome Revaud \and Diane Larlus}
\institute{Computer Vision Group, Xerox Research Center Europe\\
  \email{firstname.lastname@xrce.xerox.com}
}

\date{Received: date / Accepted: date}

\maketitle
\begin{abstract}
  While deep learning has become a key ingredient in the top performing
  methods for many computer vision tasks, it has failed so far to bring similar improvements to instance-level image retrieval. In this article,
  we argue that reasons for the underwhelming results of deep methods
  on image retrieval are threefold: i) noisy training data, ii)
  inappropriate deep architecture, and iii) suboptimal training
  procedure. We address all three issues.
  First, we leverage a large-scale but noisy landmark dataset and develop an automatic cleaning method that produces a suitable training set for deep retrieval.
  Second, we build on the recent R-MAC descriptor, show that it can be interpreted as a deep and differentiable architecture, and present improvements to enhance it.
  Last, we train this network with a siamese architecture that combines three streams with a triplet loss. 
  At the end of the training process, the proposed architecture produces a global image representation in a single forward pass that is well suited for image retrieval.
  Extensive experiments show that our approach significantly outperforms previous retrieval approaches, including state-of-the-art methods based on costly local descriptor
  indexing and spatial verification.
  On Oxford 5k, Paris 6k and Holidays, we respectively report 94.7, 96.6, and 94.8 mean average precision.
  Our representations can also be heavily compressed using product quantization with little loss in accuracy.
  To ensure the reproducibility of our research we have also released the clean annotations of the dataset and our pretrained models:
  \url{http://www.xrce.xerox.com/Deep-Image-Retrieval}.
\end{abstract}

\section{Introduction}
\textit{Instance-level image retrieval} is a visual search task that aims at, given a query image, retrieving all images that contain the same object instance as the query within a potentially very large database of images. Image retrieval and other related visual search tasks have a wide range of applications,  \eg, reverse image search on the web or organization of personal photo collections. Image retrieval has also been seen as a crucial component for data-driven methods that use visual search to transfer annotations associated with the retrieved images to the query image~\citep{tiny}. This has proved useful for annotations as diverse as image-level tags \citep{Makadia:08}, GPS coordinates \citep{Hays:08}, or prominent object location \citep{Rodriguez2015}.

Deep learning, and particularly deep convolutional neural networks (CNN), have become an extremely powerful tool in computer vision. 
After \cite{Krizhevsky2012} achieved the first place on the ImageNet classification and localization challenges in 2012 \citep{ILSVRC15} using a convolutional neural network, deep learning-based methods have significantly improved the state of the art in other tasks such as object detection~\citep{Girshick2014} and semantic segmentation \citep{Long2015}. Recently, they have also shined in other semantic tasks such as image captioning \citep{frome13devise,karpathy14deep} and visual question answering~\citep{Antol2015VQA}.
However, deep learning has been less successful so far in instance-level image retrieval. On most retrieval benchmarks, deep methods perform worse than conventional methods that rely on local descriptor matching and reranking with elaborate spatial verification~\citep{Mikulik2010,Tolias2015,Tolias2015b,Xinchao2015}.

Most of the deep retrieval methods use networks as local feature extractors, leveraging models pretrained on large image classification datasets such as ImageNet \citep{Deng2009}, 
and only focus on designing image representations suited for image retrieval on top of these features. 
Contributions have been made to allow deep architectures to accurately represent input images of different sizes and aspect ratios~\citep{Babenko2015,Kalantidis2016,Tolias2016} or to address the lack of geometric invariance of  CNN-based features~\citep{Gong2014,Razavian2014}.
Here, we argue that one of the main reasons that prevented previous retrieval methods based on deep architectures to perform well is their lack of supervised learning for the specific task of instance-level image retrieval.

In this work, we focus on the problem of \textit{learning representations that are well suited for the retrieval task}. Unlike features that are learned to distinguish between different semantic categories, and hence that are supposedly robust to intraclass variability, here we are interested in distinguishing between particular objects, even if they belong to the same semantic class. We propose a solution that combines a representation tailored for the retrieval task together with a training procedure that explicitly targets retrieval. 

For the representation, we build on the regional maximum activations of convolutions (R-MAC) descriptor \citep{Tolias2016}. 
This approach computes CNN-based descriptors of several image regions at different scales that are sum-aggregated into a compact feature vector of fixed length, and is therefore moderately robust to scale and translation.
An advantage of this method is that it can encode images at high resolutions and without distorting their aspect ratio.
However, in its original form, the R-MAC descriptor uses a CNN pretrained on ImageNet, which we believe is sub-optimal.
In our work, we note that all the steps of the R-MAC pipeline can be integrated in a single CNN and we propose to learn its weights in an end-to-end manner, as all the steps involved in its computation are differentiable. 

For the training procedure, we use a siamese network that combines three streams with a triplet loss and that explicitly optimizes the weights of our network to produce representations well suited for a retrieval task.
Furthermore, we also propose to learn the pooling mechanism of the R-MAC descriptor.
In the original architecture of \cite{Tolias2016}, a rigid grid determines the location of regions that are pooled to produce the final image-level descriptor. Here we propose to explicitly learn how to choose these regions given the image content using a region proposal network.
The training procedure results in a novel architecture that is able to encode one image into a compact fixed-length vector in a single forward pass. Representations of different images can be then compared using the dot-product.
Finally, we propose a way to encode information at different resolutions into a single descriptor. Input images are first resized at different scales and their representations are then combined, yielding a multi-resolution descriptor that significantly improves the results.

Learning the weights of our representation requires appropriate training data. To that aim we leverage the public Landmarks dataset of \cite{Babenko2014}, which is well aligned with the standard instance-level retrieval benchmarks as shown by \cite{Babenko2014}, and where images were retrieved by querying image search engines with the name of several famous landmarks. We propose a cleaning process for this dataset that automatically discards the large amount of mislabeled images and estimates the landmark location without the need of further annotations or manual intervention.

An extensive experimental study on four standard image retrieval benchmarks quantitatively evaluates the impact of each of our contributions. 
We also show the effect of combining our representation with query expansion and database-side feature augmentation, and the impact of compression with product quantization.
In the end, we obtain results that largely outperform the state of the art on all datasets, not only compared to methods that use one global representation per image, but also against much more costly methods that, unlike our proposed method, require to perform a subsequent matching stage or geometrical verification.

The rest of the paper is organized as follows. Section \ref{sec:rw} discusses related work. Section \ref{sec:cleaning} describes the cleaning procedure that leads to a suitable training set. Section \ref{sec:part1} describes the training procedure while Section \ref{sec:part2} proposes several improvements to our deep architecture. Section \ref{sec:sota} describes the final pipeline and compares it with the state of the art. Finally, Section \ref{sec:conclusions} concludes the paper.

This article extends our previous work \citep{gordo2016deep} in the following manner: we consider residual network architectures as an alternative when constructing our global descriptor (and their very deep nature requires to adjust our training procedure, see Section~\ref{sec:deeper}).
 We build a multi-resolution version of the descriptor to cope with scale changes between query and database images (Section \ref{sub:multires}).
 We propose to combine our method with database-side feature augmentation to significantly improve the retrieval accuracy with no extra cost at testing time (Section~\ref{sub:dba}). 
We evaluate the effect of compression in our representation, both with PCA and with product quantization (Section~\ref{sub:part3xps}).
These new contributions lead to significantly improved results.
Furthermore, we also show qualitative results illustrating the impact of learning in the model activations. 

\section{Related work on image retrieval}
\label{sec:rw}

This section gives an overview of some of the key papers that have contributed to instance-level image retrieval.

\subsection{Conventional image retrieval} 
Early techniques for instance-level retrieval such as the ones of
\cite{Sivic2003}, \cite{Nister2006}, and \cite{Philbin2007} rely on bag-of-features representations,
large vocabularies, and inverted files. Numerous methods that better
approximate the matching of the descriptors have been proposed, see
for instance the works of \cite{Jegou2008,Jegou2010,Mikulik2013,Tolias2015}.  An advantage of these
techniques is that spatial verification can be employed to rerank a
shortlist of results \citep{Philbin2007,Perdoch2009}, yielding a
large improvement despite a significant cost.

Concurrently, methods that aggregate local patches to build a global image representation have been considered. Encoding techniques, such as the
Fisher Vector~\citep{Perronnin2007,Perronnin2010} or the VLAD descriptor \citep{Jegou2010aggregating} have been used for example by \cite{Perronnin2010, Gordo2012, Jegou2012,Radenovic2015}.
All these methods can be combined with postprocessing techniques such as query expansion \citep{Chum2007,Chum2011,Arandjelovic2012three}.
Some works also suggest to compress the descriptors to improve the storage requirements and retrieval efficiency at the cost of reduced accuracy. 
Although the most common approach is to use unsupervised compression through PCA or product quantization \citep{Perronnin2010,Jegou2012,Radenovic2015}, supervised dimensionality reduction approaches are also possible \citep{Gordo2012}.

\subsection{CNN-based retrieval} In the seminal work of \cite{Krizhevsky2012}, the activations of a CNN trained for ImageNet classification were used as image features for an instance-level retrieval task, although this was only evaluated in qualitative terms.
Soon after, these off-the-shelf CNN features were evaluated quantitatively by \cite{Razavian2014}.
 Several improvements were proposed
to overcome their lack of robustness to scaling, cropping and image clutter. The method of \cite{Razavian2014} performs
region cross-matching and accumulates the maximum similarity per query region while the one of \cite{Babenko2015} applies sum-pooling
to whitened region descriptors. 
\cite{Kalantidis2016} extended the work of \cite{Babenko2015} by allowing cross-dimensional
weighting and aggregation of neural codes. Other approaches proposed hybrid models also involving an encoding technique such
as \cite{Perronnin2015} that used the FV or \cite{Gong2014} and \cite{Paulin2015} that considered VLAD.
Although these methods outperform standard
global descriptors, their performance is significantly below the state of the art of conventional methods.

\cite{Tolias2016} proposed to aggregate the activation features of a CNN in a fixed layout of spatial regions.
 The method uses a pretrained, fully convolutional CNN to extract local features of images without distorting their aspect ratio and independently of their size,
 and aggregates these local features into a global representation using normalizations known to work well for image retrieval \citep{Jegou2012}.
 The result is the R-MAC descriptor, a fixed-length vector representation of the image that, when combined with query expansion, achieves results close to the state of the art. 
Our work draws inspiration from the R-MAC pipeline, but learns the model weights in an end-to-end manner.

\subsection{Finetuning for retrieval}
The use of off-the-shelf features from models trained for classification on ImageNet may not be the optimal choice for instance-level retrieval tasks due to the models being trained to achieve intraclass generalization. Instead of using pretrained models as a feature extractor, a few methods have proposed to explicitly learn weights more suited for the retrieval task. The work of \cite{Babenko2014} showed that models pretrained on ImageNet for object classification could be improved by finetuning them on an external set of Landmarks images, even when using a classification loss. 

A preliminary version of our work \citep{gordo2016deep}, together with a concurrent work \citep{Radenovic2016}, confirmed that finetuning the pretrained models for retrieval can bring a significant improvement, but demonstrated that even more crucial are the combination of i) a good image representation and ii) a ranking loss -- as opposed to the classification loss used by \cite{Babenko2014}. The recent NetVLAD by \cite{Arandjelovic2016} also highlights the importance of learning to rank.

\begin{figure*}[t!]
\begin{centering}
\resizebox{\linewidth}{!}{\includegraphics[height=0.27\linewidth]{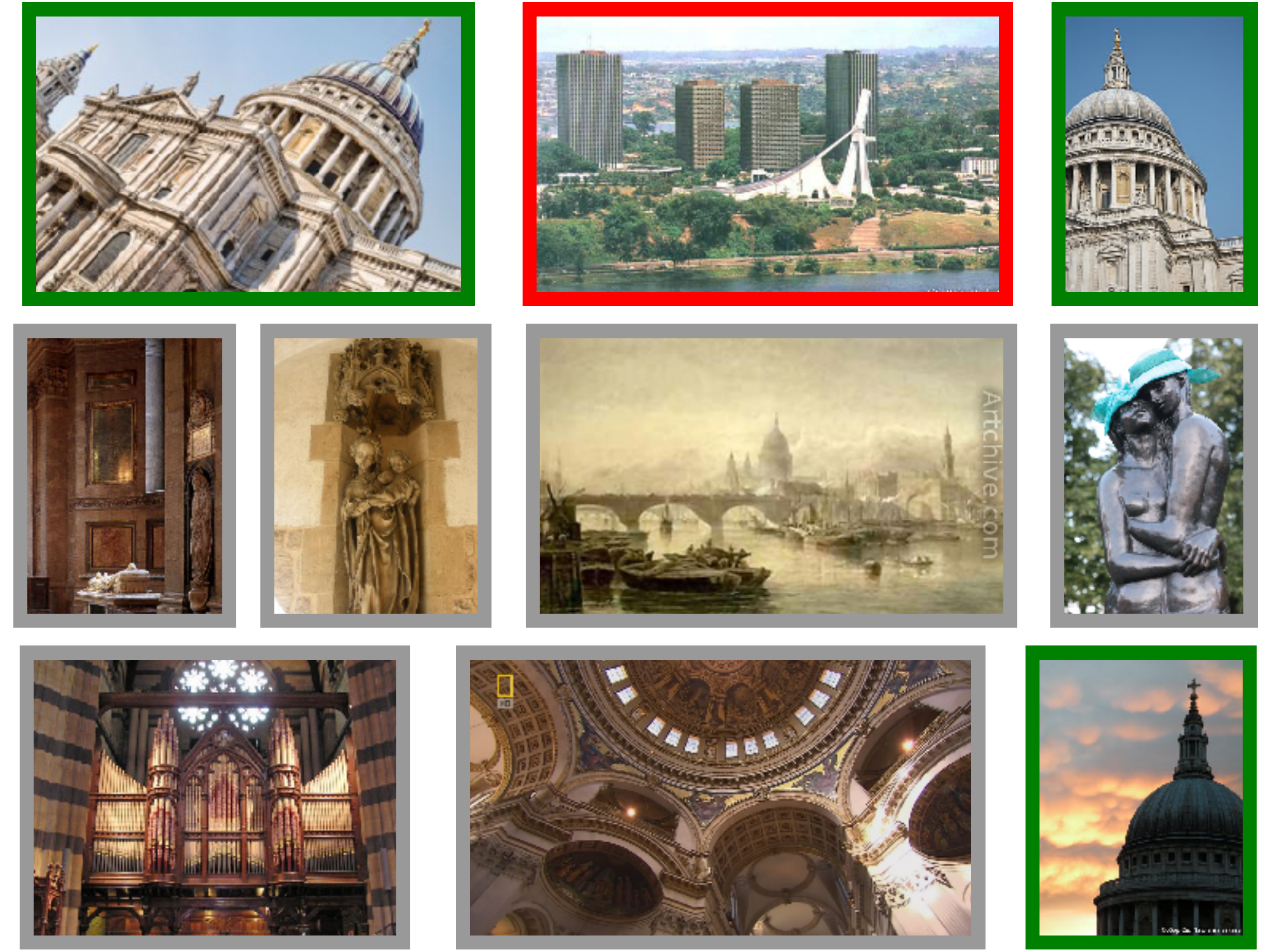}\hspace{12mm}\includegraphics[bb=30bp 150bp 930bp 530bp,clip,height=0.27\linewidth]{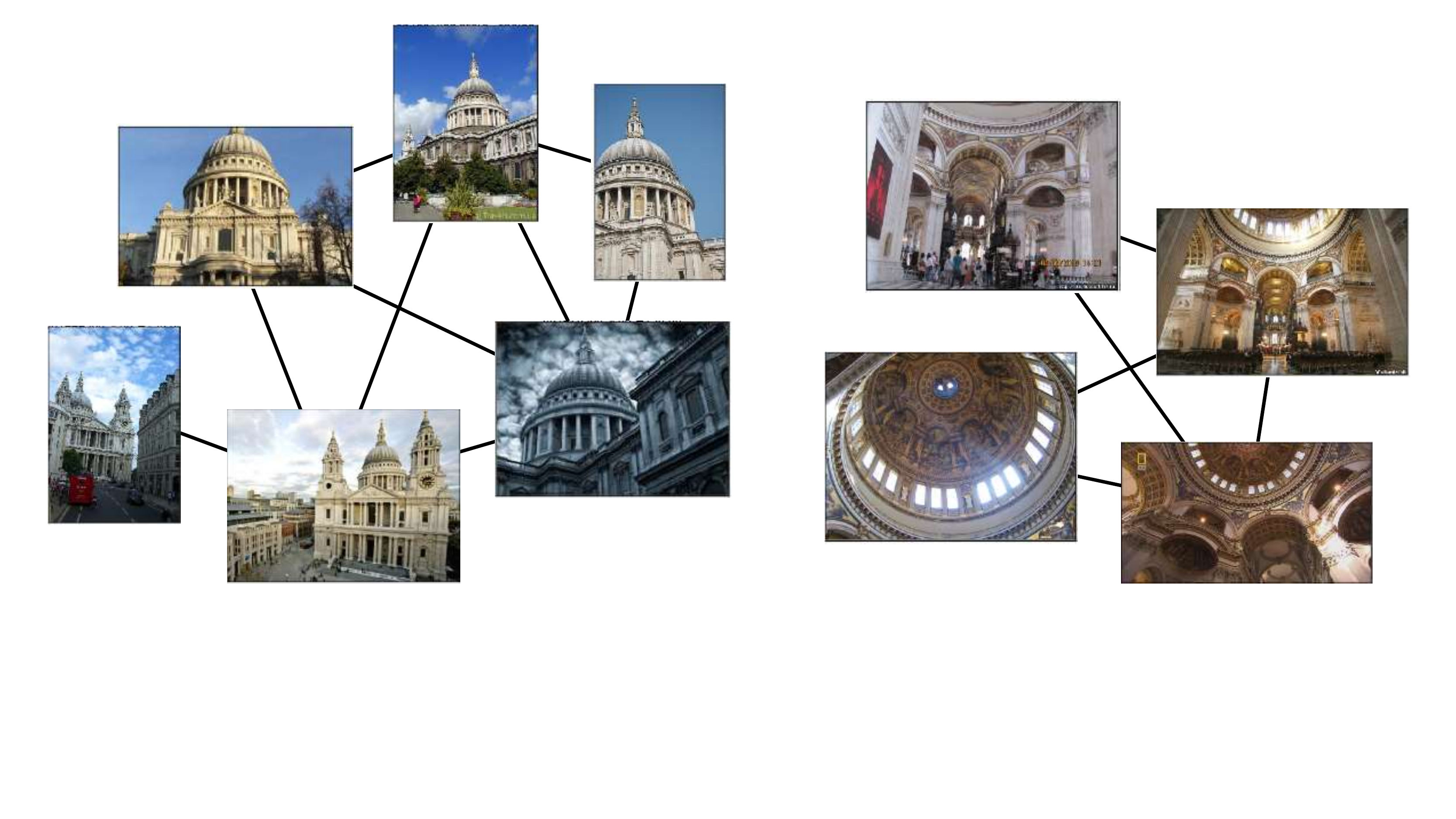}}
\par\end{centering}
\caption{\label{fig:cc}\textbf{Left}: random images from the ``St Paul's Cathedral''
landmark. Green, gray and red borders respectively denote prototypical, non-prototypical, and incorrect images.
\textbf{Right}: excerpt of the two largest connected components of the pairwise matching graph (corresponding
to outside and inside pictures of the cathedral).}
\end{figure*}

\section{Leveraging large-scale noisy data}
\label{sec:cleaning}

To learn an informative and efficient representation for
instance-level retrieval, we need the appropriate dataset. This
section describes how we leveraged and
automatically cleaned an existing dataset to obtain the
characteristics we need for training our models.

We leverage the \textbf{Landmarks} dataset~\citep{Babenko2014}, a large-scale image dataset that contains approximately $214$k images of $672$ famous landmark sites.  Its images were collected through
textual queries in an image search engine without thorough verification.  As a consequence, they comprise a large variety of profiles: general views of the site, close-ups of details like statues
or paintings, with all intermediate cases as well, but also site map pictures, artistic drawings, or even completely unrelated images, see Fig.~\ref{fig:cc}.

We could only download a subset of all images due to broken URLs. We removed classes with too few
images. We also meticulously removed all classes having an overlap
with the Oxford 5k, Paris 6k, and Holidays datasets, on which we
experiment, see Section~\ref{sec:exp-siamese}.  We obtained a set of about
192,000 images divided into $586$ landmarks.  We refer to this set as
\textbf{Landmarks-full}.  For our experiments, we use 168,882 images
for the actual finetuning, and the 20,668 remaining ones to validate
parameters.

\paragraph{Cleaning the Landmarks dataset.} The
Landmarks dataset presents a non-negligible amount of unrelated images (Fig.~\ref{fig:cc}). While this could be allowed 
in certain frameworks (\eg for classification, typically networks can
accommodate during training for this diversity and even for noise), in
some scenarios we need to learn our representations with images of
the \emph{same} particular object or scene. In this case, variability comes
from different viewing scales, angles, lighting conditions and image
clutter. We preprocess the Landmarks dataset to achieve this as
follows.


We first run a strong image
matching baseline within the images of each landmark class.  We compare each pair of images using invariant keypoint
matching and spatial verification~\citep{Lowe2004}. We use the SIFT and Hessian-Affine keypoint detectors
\citep{Lowe2004,Mikolajczyk2004} and match keypoints using the first-to-second neighbor ratio rule~\citep{Lowe2004}. This is known
to outperform approaches based on descriptor quantization~\citep{Philbin2010}. Afterwards, we verify all pairwise matches with an
affine transformation model as proposed by~\cite{Philbin2007}. This heavy procedure is affordable as it is performed offline, only once at training time, and on a class per class basis. 

Without loss of generality, we describe the rest of the cleaning procedure for a single landmark class. Once we have
obtained a set of pairwise scores between all image pairs, we construct a graph whose nodes are the images and edges are
pairwise matches.  We prune all edges which have a low score.
Then we extract the connected components of the graph.  They correspond to different
profiles of a landmark; see Fig.~\ref{fig:cc} that shows the two largest connected components for St Paul's Cathedral.
Finally we retain only the largest connected component and discard the others to ensure that all images inside a class are visually related.
This cleaning process leaves about 49,000 images (divided in 42,410 training and 6,382 validation images) still belonging to one of the 586 landmarks, referred to as \textbf{Landmarks-clean}. 
The cleaning process took approximately one week on a 32-core server, parallelizing over classes.

\begin{figure*}[ht!]
\begin{centering}
\resizebox{0.85\linewidth}{!}{\includegraphics[bb=210bp 80bp 770bp 460bp,clip,height=0.3\linewidth]{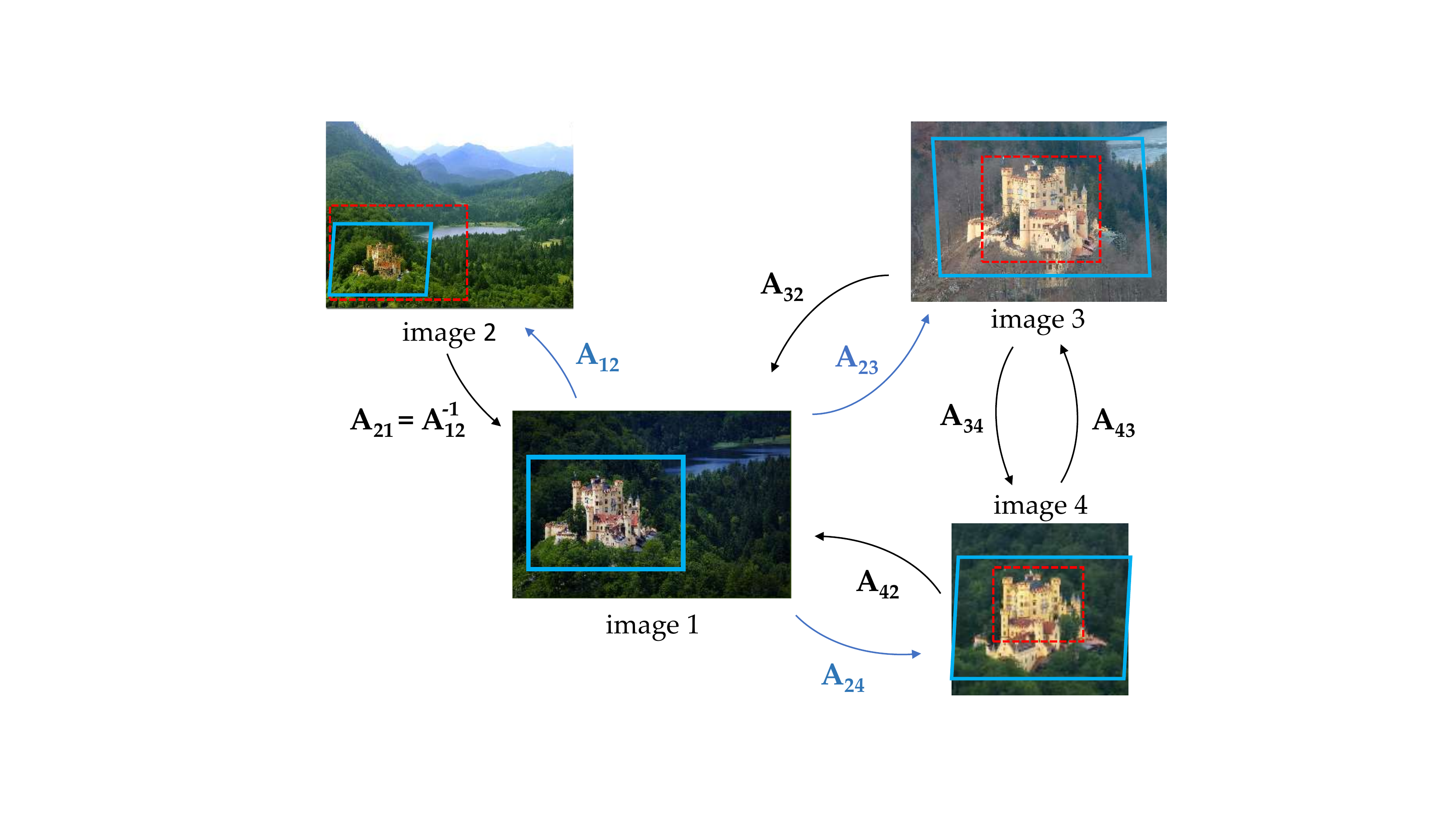}
\hspace{10mm}\includegraphics[bb=25bp 40bp 560bp 360bp,clip,height=0.3\linewidth]{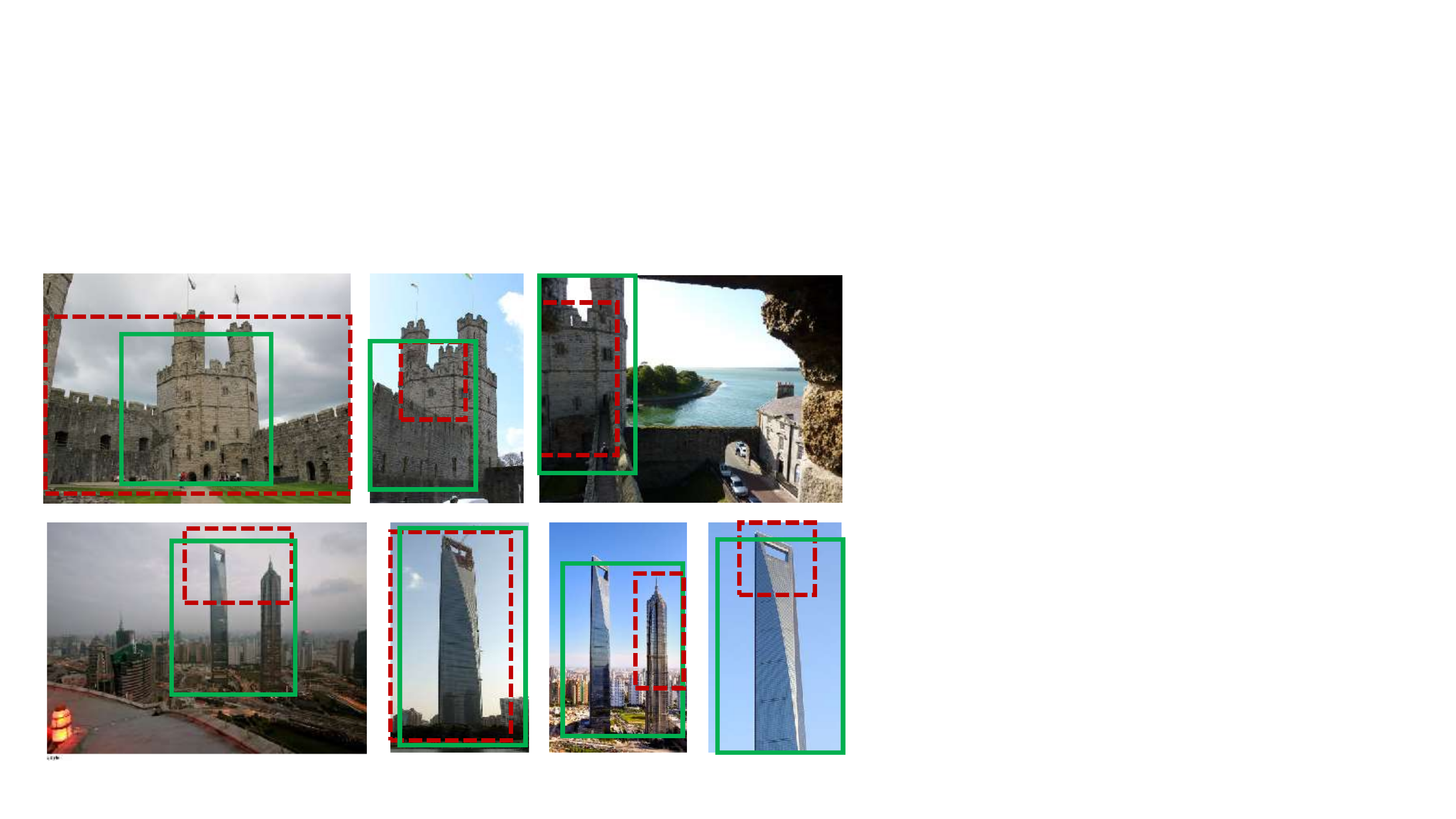}}
\par\end{centering}
\caption{\label{fig:diffusion}\textbf{Left}: the bounding box from image 1 is projected
into its graph neighbors using the affine transformations (blue rectangles). 
The current bounding box estimates (dotted red rectangles) are then updated accordingly.
The diffusion process repeats through all edges until convergence.
\textbf{Right}: initial (dotted red box) and final (solid green box) estimates.}
\end{figure*}

{\paragraph{Bounding box estimation.}
In one of our experiments, we replace the uniform sampling of regions in
the R-MAC descriptor by a learned region of interest (ROI) selector (Section~\ref{sec:proposal}).
This selector is trained using bounding box annotations that we automatically estimate for all landmark images.
To that aim we leverage the data obtained during the cleaning step.
The position of verified keypoint matches is a meaningful cue since the object of interest is consistently
visible across the landmark's pictures, whereas distractor backgrounds or foreground objects are varying and hence
unmatched.}

{We denote the connected component from each landmark as a graph $\mathcal{S}=\left\{
\mathcal{V}_{\mathcal{S}},\mathcal{E}_{\mathcal{S}}\right\} $.  Each pair of connected images
$(i,j)\in\mathcal{E}_{\mathcal{S}}$ corresponds to a set of verified keypoint matches and an affine
transformation $A_{ij}$.  We first define an initial bounding box in both images $i$ and $j$, denoted by $B_{i}$ and
$B_{j}$, as the minimum rectangle enclosing all matched keypoints. Note that a single image can be involved in many
different pairs. In this case, the initial bounding box is the geometric median of all boxes, efficiently computed
as in~\cite{GeoMedian2004}. Then, we run a diffusion process, illustrated in Fig.~\ref{fig:diffusion}, in which for a
pair $(i,j)$ we predict the bounding box $B_{j}$ using $B_{i}$ and the affine transform $A_{ij}$ (and conversely).  At
each iteration, bounding boxes are updated as: $B_{j}'=(\alpha-1)B_{j}+\alpha A_{ij}B_{i}$, where $\alpha$ is a small
update step (we set $\alpha=0.1$ in our experiments). Again, the multiple updates for a single image are merged using
geometric median, which is robust against poorly estimated affine transformations.  This process iterates until
convergence. As can be seen in Fig.~\ref{fig:diffusion}, the locations of the bounding boxes are improved as well as 
their consistency across images.}
We are making the list of Landmarks-clean images and the estimated bounding boxes available. 

Next we leverage our cleaned dataset to learn powerful image representations tailored for image retrieval.

\section{Learning to rank: an end-to-end approach}
\label{sec:part1}

\begin{figure*}[t!]
\includegraphics[width=1\linewidth]{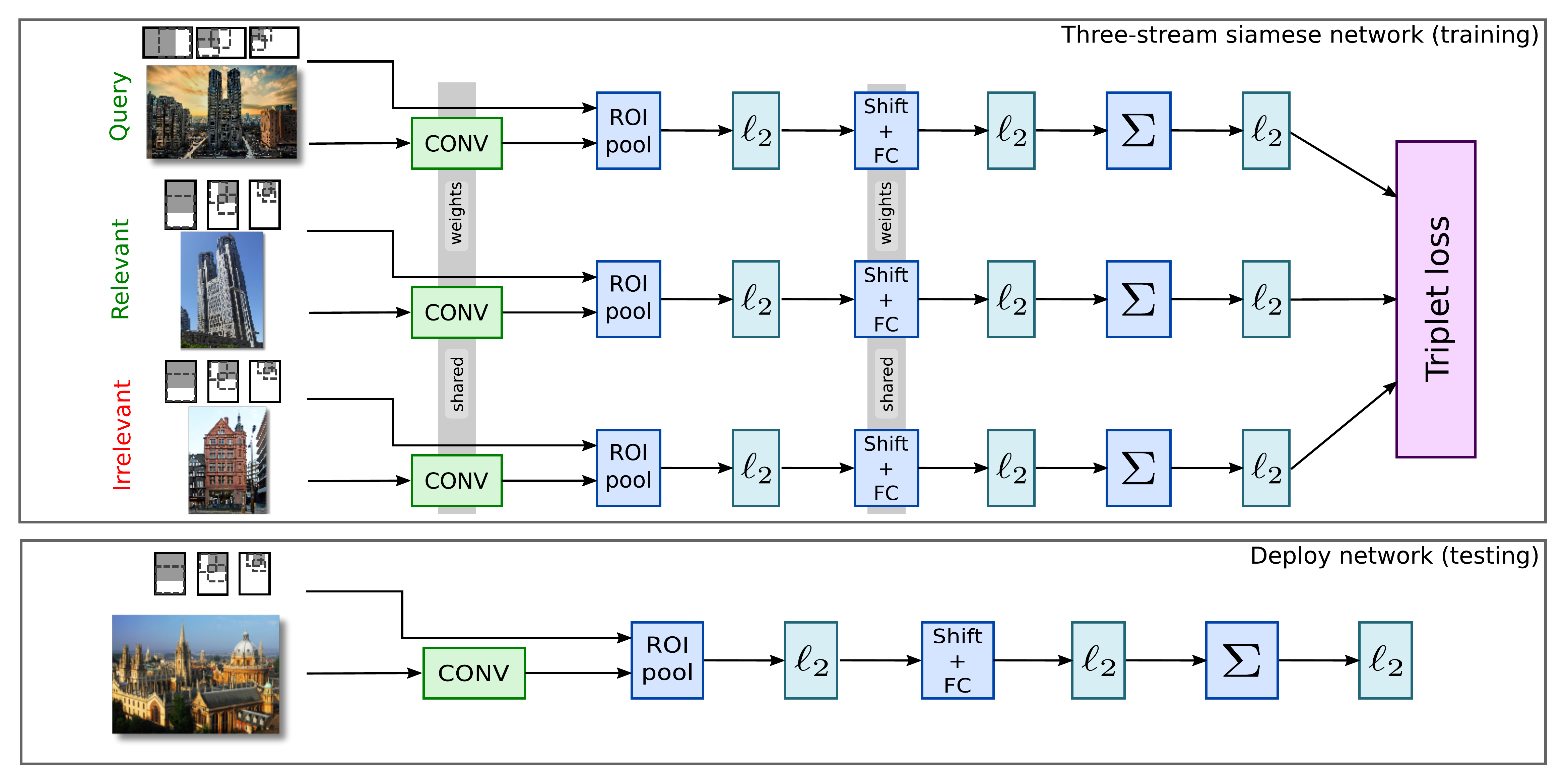}
\caption{\textbf{Proposed siamese network.} At training time, image triplets are
sampled and simultaneously considered by a \textit{triplet-loss} that is well-suited for the task (top).
At test time, the query image is fed to the learned architecture to
efficiently produce a \textit{compact global image representation} that can be compared with the dataset image representations with a
dot-product (bottom). \label{fig:siamese}}
\end{figure*}

\label{sec:method}

This section first revisits the R-MAC representation of \cite{Tolias2016} in Section \ref{sec:r-mac} and shows that, despite its handcrafted nature, all the operations involved in it can be integrated into a single CNN that computes the R-MAC representation in one single forward pass. 
More importantly, all of its components consist of differentiable operations, and therefore, given training data and an appropriate loss, one can learn the optimal weights of the architecture in an end-to-end manner.
To that aim we leverage a three-stream siamese network with a triplet ranking loss (Section~\ref{sec:learning}).
Then we discuss the practical details that allow this architecture to scale to deep networks with large memory needs (Section~\ref{sec:deeper}).
Finally, we experimentally validate the gain obtained by the proposed training strategy in terms of accuracy in standard benchmarks (Section~\ref{sec:exp-siamese}).

\subsection{The R-MAC baseline}
\label{sec:r-mac}

The R-MAC descriptor, recently introduced by \cite{Tolias2016}, is
a global image representation that is particularly well-suited for image retrieval.
At its core, it uses a ``fully convolutional'' CNN as a powerful local feature extractor that works independently of the image size and that extracts local features without distorting the aspect ratio of the original image.
The original work of \cite{Tolias2016} uses both AlexNet \citep{Krizhevsky2012} and VGG16 \citep{Simonyan2015verydeep} network architectures, with models pretrained on the ImageNet dataset, but other network architectures such as residual networks \citep{He2016} can also be used.
These local features are then max-pooled across several multi-scale overlapping regions, obtained from a rigid grid covering the image, similar in spirit to spatial pyramids, producing a single feature vector per region.
These region-level features are independently $\ell_2$-normalized, whitened with PCA, and $\ell_2$-normalized again, a normalization pipeline known to work well for image retrieval \citep{Jegou2012}.
Finally, region descriptors are sum-aggregated and $\ell_2$-normalized once again.
The obtained global image representation is a compact vector whose size (typically $256$ to $2k$ dimensions, depending on the network architecture) is independent of the size of the image and of the number of regions.
Note that the region pooling is different from a spatial pyramid: the latter concatenates the region descriptors, while the former sum-aggregates them.
Comparing the R-MAC vectors of two images with a dot-product can then be interpreted as a weighted many-to-many region matching, where the weights depend on the norm of the aggregated region descriptors.



\subsection{Learning to retrieve}
\label{sec:learning}

One key aspect of the R-MAC pipeline is that all of its components are differentiable operations. More precisely, the multi-scale spatial pooling in different regions is equivalent to the Region of Interest (ROI) pooling from \cite{He2014} using a fixed rigid grid, which is differentiable
as shown in the context of detection \citep{Girshick2015}.
The PCA projection can be seen as a combination of a shifting (for the mean centering) and a fully connected (FC) layer (for the projection with the eigenvectors), with weights that can be learned. The sum-aggregation of the different regions and the $\ell_2$-normalization are also differentiable.
Therefore, one can implement a network architecture that, given an image and the precomputed coordinates of its regions, directly produces a representation equivalent to the R-MAC pipeline.
As all the components are differentiable, one can backpropagate through the network architecture to learn the optimal network weights, namely the weights of the convolutions and of the shifting and fully-connected layers that replace the PCA.

\paragraph{Learning with a classification loss.}
One can easily finetune a standard classification architecture (\eg VGG16) on the Landmarks dataset using a cross-entropy loss, as previously done by \cite{Babenko2014}, and then use the improved convolutional filters as the feature extractor of the R-MAC pipeline, instead of using the original weights. We use this approach as our training baseline, and note that it has important issues.
First, it does not learn directly the task to address, retrieval, but a proxy, classification.
Second, it does not leverage the R-MAC architecture, as it learns on the original classification architecture, using low-resolution square crops.
The convolutional weights are used together with the R-MAC architecture only after the training has finished.
In our experiments we show how this naive finetuning method already outperforms the baseline approach significantly, but does not match the accuracy obtained by training using the appropriate architecture and loss.

\paragraph{Learning with a ranking loss.}
In our work we propose to consider a ranking loss based on image triplets. The goal is to explicitly enforce that, given a triplet composed of a query image, a relevant element to the query, and an irrelevant one, the R-MAC representation of the relevant image is closer to the representation of the query than the representation of the irrelevant one. 

We design a three-stream siamese network architecture where the image representation produced by each of the three streams are jointly considered by the loss. This architecture is illustrated in Fig.~\ref{fig:siamese}. The weights of the convolutional filters and of the fully-connected layer are shared between the  streams as their size is independent of the size of the images. This means that the siamese architecture can process images of any sizes and aspect ratios, and we can train the network using images at the same (high) resolution that is used at test time. 

Siamese networks have performed well for metric learning \citep{Song2015}, dimensionality reduction \citep{Hadsell2006}, learning image descriptors \citep{Serra2015}, and performing face identification \citep{Chopra2005,Hu2014,Sun2014}. Recently triplet networks (i.e. three-stream siamese networks) have been considered for metric learning \citep{Hoffer2015,Wang2014} and face identification \citep{SchroffK2015}. 

We use the following ranking loss. Let $I_q$ be a query image with R-MAC descriptor $q$, $I^+$ be a relevant image with descriptor $d^+$, and $I^-$ be an irrelevant image with descriptor $d^-$.
We define the ranking triplet loss as
\begin{equation}\label{eq:loss}
L(I_q,I^+,I^-) = \frac{1}{2} \max (0, m + \|q-d^+\|^2 - \|q-d^-\|^2),
\end{equation}
where $m$ is a scalar that controls the margin. 
Given a triplet that produces a non-zero loss, the sub-gradients are given by:
\begin{equation}\label{eq:grad}
    \centering
    \frac{\partial L}{\partial q} = d^- - d^+, \quad 
    \frac{\partial L}{\partial d^+} = d^+ - q, \quad 
    \frac{\partial L}{\partial d^-} = q - d^-. 
\end{equation}

The sub-gradients are backpropagated through the three streams of the
network, and the convolutional layers together with the ``PCA'' layers -- the shifting and the fully connected layer -- get updated.
This approach directly optimizes a ranking objective.

\subsection{Practical considerations}
\label{sec:deeper}
When learning with a ranking loss, one should pay attention to certain practical considerations.
The first one is the sampling of the triplets, as sampling them randomly will, most of the time, yield triplets that incur no loss and therefore do not improve the model.
To ensure that the sampled triplets are useful, we first select randomly $N$ training samples, extract their features with the current model, and compute all possible triplets and their losses, which is fast once the features have been extracted. All the triplets that incur a loss are preselected as good candidates. Triplets can then be sampled from that set of good candidates, with a bias towards hard triplets, \ie triplets that produce a high loss.
In practice this is achieved by randomly sampling one of the $N$ images with a uniform distribution and then randomly choosing one of the $25$ triplets with the largest loss that involve that particular image as a query. 
Note that, in theory, one should recompute the set of good candidates every time the model gets updated, which is very time consuming. In practice, we assume that most of the hard triplets for a given model will remain hard even if the model gets updated a few times, and therefore we only update the set of good candidates after the model has been updated $k$ times. We used  $N = 5,000$ samples and $k=64$ iterations with a batch size of $64$ triplets per iteration in our experiments.

The second consideration is the amount of memory required during training, as we train with large images (larger side resized to 800 pixels) and with three streams at the same time. When using the VGG16 architecture, we could only fit one triplet in memory at a time on an M40 GPU with 12 Gb of memory. To perform updates with a batch of effective size $bs$ larger than one, we sequentially compute and aggregate the gradients of the loss with respect to the parameters of the network for every triplet, and only perform the actual update every $bs$ triplets, with $bs=64$.

When using a larger network such as ResNet101, the situation becomes more complex, as we do not have enough memory to process even one single triplet. Instead of reducing the image size, which would result in a loss of detail, we propose an alternative approach detailed in Algorithm \ref{alg:sst}. This approach allows us to process the streams of a triplet sequentially using one single stream instead of all of them simultaneously. This yields exactly the same gradients but trades some computational efficiency due to recomputations (about a $25\%$ overhead) for very significant memory reduction (only one third of the memory is required, from 23 Gb down to 7.5 Gb). This allows one to train the model using very deep architectures without reducing the size of the training images.

\begin{algorithm}
\caption{Memory efficient model update}
\begin{algorithmic}[1]
\Procedure{Process Triplet}{}
\State $Q$: Query image
\State $I^+$: Relevant image
\State $I^-$: Irrelevant image
\BState \emph{Main}:
\State Compute feature representation of $Q$: $q$ 
\State Compute feature representation of $I^+$: $d^+$ 
\Statex \textit{\color{gray}~~~~~~/Overwrites results needed to backpropagate}
\Statex \textit{\color{gray}~~~~~~~~the loss with respect to $q$/}
\State Compute feature representation of $I^-$: $d^-$ 
\Statex \textit{\color{gray}~~~~~~/Overwrites results needed to backpropagate the loss}
\Statex \textit{\color{gray}~~~~~~~~with respect to $d^+$/}
\State Compute loss as in Equation \eqref{eq:loss}
\State Compute gradients with respect to $q$, $d^+$, and $d^-$ as
\Statex {~~~~ in Equation \eqref{eq:grad}}
\State Backpropagate the loss with respect to $d^-$
\State Recompute $q$
\Statex \textit{\color{gray}~~~~~~/recomputing is needed to obtain the necessary}
\Statex \textit{\color{gray}~~~~~~statistics to backpropagate/}
\State Backpropagate the loss with respect to $q$
\State Recompute $d^+$ 
\Statex \textit{\color{gray}~~~~~~/recomputing is needed to obtain the necessary}
\Statex \textit{\color{gray}~~~~~~statistics to backpropagate/}
\State Backpropagate the loss with respect to $d^+$
\EndProcedure
\end{algorithmic}
\label{alg:sst}
\end{algorithm}




\subsection{Experiments}
\label{sec:exp-siamese}

In this section we study the impact of learning the weights for different setups and architectures. In all these experiments we assume that the descriptor extracts region following the standard R-MAC strategy, i.e. following a predefined rigid grid. 

\subsubsection{Experimental details}
We test our approach on four standard datasets: the \textit{Oxford 5k} building dataset \citep{Philbin2007}, the \textit{Paris 6k} dataset
\citep{Philbin2008}, the INRIA \textit{Holidays} dataset~\citep{Jegou2008}, and the University of Kentucky Benchmark \textit{(UKB)} dataset~\citep{Nister2006}.
We use the standard evaluation protocols, \ie recall@4 for UKB and mean average precision (mAP) for the rest.
As is standard practice, in Oxford and Paris one uses only the annotated region of interest of the query, while for Holidays and UKB one uses the whole query image.
Furthermore, the query image is removed from the dataset when evaluating on Holidays, but not on Oxford, Paris, and UKB.
Following most CNN-based methods, we manually correct the orientation of the images on the Holidays dataset and evaluate on the corrected images.
For fair comparison with methods that do not correct the orientation we also report results without correcting the orientation in our final experiments.
 
For the convolutional part of our network, we evaluate two popular architectures: VGG16  \citep{Simonyan2015verydeep} and ResNet101 \citep{He2016}. In both cases we start with the publicly available models pretrained on the ImageNet ILSVRC data. The fully-connected layer is initialized with a PCA projection, computed on the normalized per-region descriptors. 
All subsequent learning is performed on the Landmarks dataset.

To perform finetuning with classification we follow standard practice and resize the training images to multiple scales (shortest side in the $\left[256-512\right]$ range) and extract random crops of $224\times 224$ pixels.
To finetune using our proposed architecture we also augment our training data performing random crops (randomly removing up to $5\%$ of each side of the image) and then resize the resulting crop such as that the larger side is of $800$ pixels, preserving the aspect ratio. At test time, all the database images are also resized so the larger side is $800$ pixels\footnote{Note that this differs from the original setup of \cite{Tolias2016}, that resizes images to 1024 pixels, and leads to different results in Table \ref{tab:ml}. Please see \cite{gordo2016deep} for a discussion about this issue.}
All the models are trained with stochastic gradient descent (SGD)  with momentum of $0.9$, learning rate of $10^{-3}$, and weight decay of $5\cdot 10^{-5}$. We decrease the learning rate down to $10^{-4}$ on the classification finetuning once the validation error on Landmarks stops decreasing. We did not see any improvement by reducing the learning rate when learning to rank, and so we keep the learning rate at $10^{-3}$ until the end. 
The margin is set to $m=0.1$.

\subsubsection{Results}


\paragraph{Quantitative evaluation.} We report results in Table~\ref{tab:ml} for two possible choices of the convolutional part of the network: VGG16 (top) and ResNet101 (bottom).
For each architecture, we first report performance with the R-MAC baseline, whose convolutional layer weights are taken directly from the ImageNet pretrained networks and the PCA is learned on Landmarks-full.
For the learned models, weights are finetuned on Landmarks either with a classification loss (\textbf{Ft-Cls}) or with a ranking loss (\textbf{Ft-Rnk}).
For the latter, we consider either initializing the weights directly with the ImageNet pretrained network or with a warmed up model already finetuned on Landmarks using a classification loss.  

From the results reported in Table~\ref{tab:ml} we highlight the following observations.

\begin{figure}[t!]
\begin{centering}
\includegraphics[width=0.51\linewidth,trim={0.28cm 0 0.3cm 0},clip]{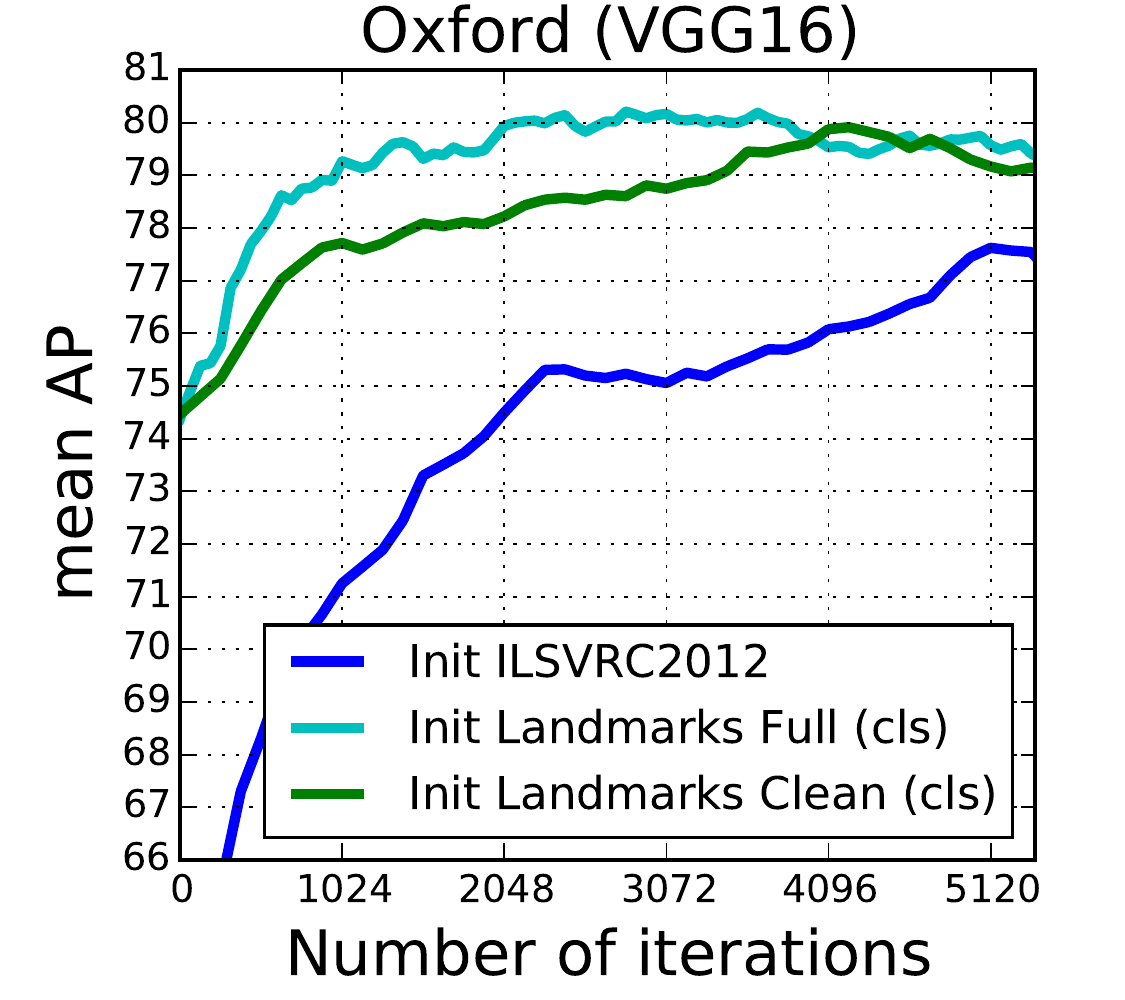}
\includegraphics[width=0.47\linewidth,trim={1.27cm 0 0.3cm 0},clip]{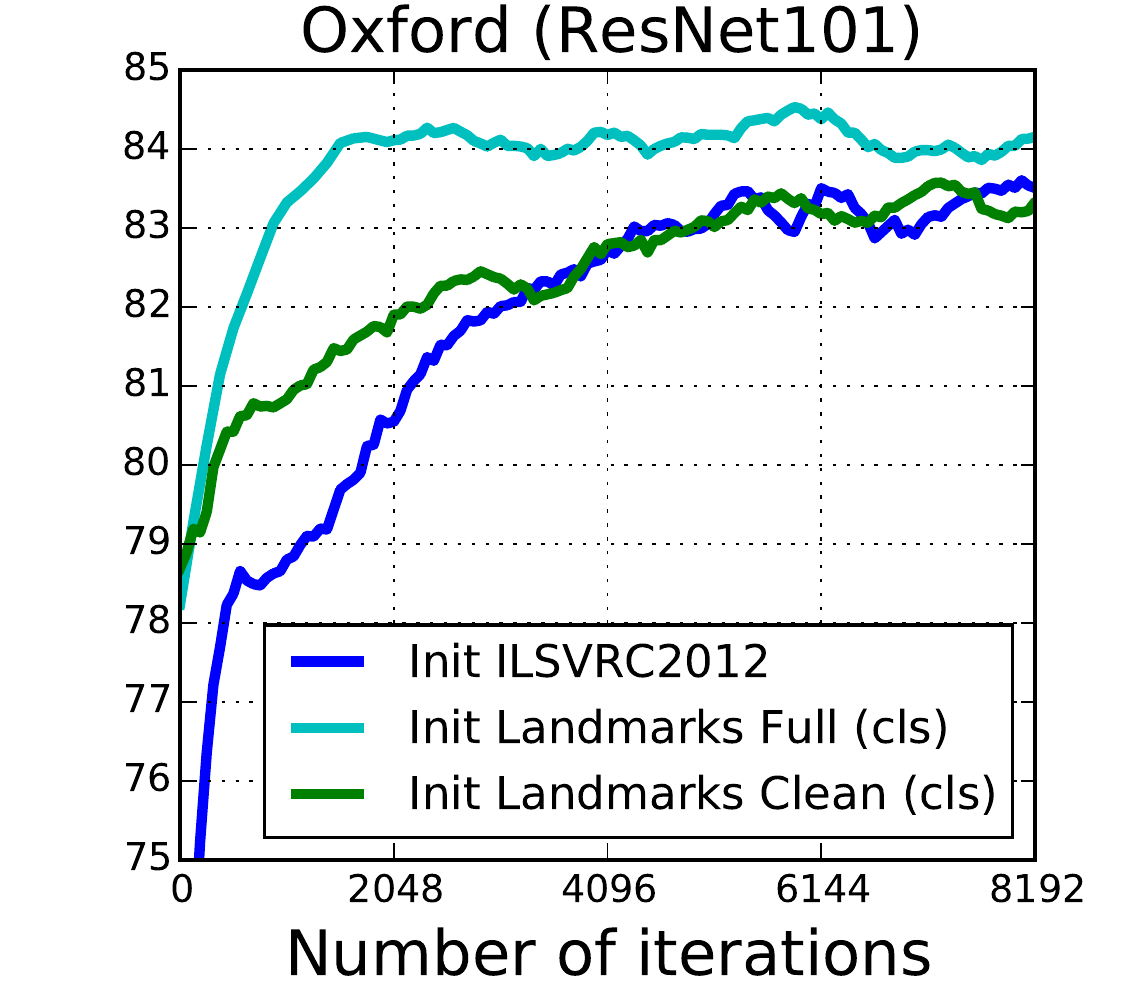}
\par\end{centering}
\caption{\label{fig:learning} Accuracy comparison of three different model initializations when finetuning the representation with a ranking loss as a function on the number of iterations (one iteration corresponds to a batch of 64 triplets).}
\end{figure}

\renewcommand\labelitemi{$\bullet$}
\begin{itemize}

\item Finetuning with a naive classification loss on a relevant dataset already brings a significant improvement over a model pretrained on ImageNet, as already observed by \cite{Babenko2014} (albeit on a different architecture) on the first three datasets.
In this case, training with Landmarks-full or training with Landmarks-clean does not make a significant difference. 

\item Finetuning our proposed architecture with a ranking loss is the best performing strategy. For the first three datasets again, it seems very beneficial to improve the weights of our model using the Landmarks dataset.
    We only report results learning the ranking with Landmarks-clean. We found this to be crucial: learning on Landmarks-full significantly worsens the accuracy of the model.
 
\item To obtain good results with VGG16 using the ranking loss we found important to warm up the network by first training it on the Landmarks dataset using a classification loss, as done in \cite{gordo2016deep}. However, this was not so important when using the more recent ResNet101 architecture: although warming up the network brings slight improvements, the final results are similar. This can also be observed in Fig.~\ref{fig:learning}, which shows the evolution of the accuracy on the Oxford dataset as training progresses for different model initializations.

\item  For the UKB dataset, the ``off-the-shelf'' R-MAC already provides state-of-the-art results, and the training slightly decreases its performance, probably because of the large domain differences (\cf Section~\ref{sec:sota} for a more detailed discussion about UKB).
 
\item As expected, the model based on ResNet101 outperforms the model based on VGG16. This gap, however, is not as significant as the improvement brought by the training. 

\end{itemize}

\begin{table*}[t!]
    \caption{Impact of learning the weights of the representation with a classification (Cls) and a ranking (Rnk) loss, either with VGG16 or ResNet101. The weights are learned either from the full Landmarks dataset (Landmarks-Full) of the clean version (Landmarks-Clean). For the ranking loss we also compare different intializations.}
\centering
\begin{tabular}{clcccc}
\toprule
Architecture & Model & Oxford 5k & Paris 6k & Holidays & UKB \\
\midrule
\multirow{6}{*}{VGG16} & ILSVRC2012 baseline & 60.3 & 79.9 & 85.8 & 3.75 \\
\cline{2-6} 
    & Ft Cls-Landmarks-Full & 74.2 & 82.5  & 87.7 & 3.65 \\
    & Ft Cls-Landmarks-Clean & 74.0 & 83.0 & 86.0 & 3.62 \\
\cline{2-6} 
    & Ft Rnk-Landmarks-Clean & 76.3 & 86.2 & 85.6 & 3.61   \\
    & Ft Cls-Landmarks-Full $\Rightarrow$  Ft Rnk-Landmarks-Clean & 79.9 & 85.9 & 87.9 & 3.59 \\
    & Ft Cls-Landmarks-Clean $\Rightarrow$  Ft Rnk-Landmarks-Clean & 79.0 & 86.9 & 86.4 & 3.55  \\
\midrule
\multirow{6}{*}{ResNet101} & ILSVRC2012 baseline & 69.4 & 85.2 & 91.4 & \textbf{3.89} \\
\cline{2-6} 
    & Ft Cls-Landmarks-Full & 77.7 & 89.4 & 93.4 & \textbf{3.89} \\
    & Ft Cls-Landmarks-Clean  &  78.5  & 88.2 & 93.0 & 3.86  \\
\cline{2-6} 
    & Ft Rnk-Landmarks-Clean  & 83.4 & 92.8 & 93.7 & 3.85\\    
    & Ft Cls-Landmarks-Full $\Rightarrow$  Ft Rnk-Landmarks-Clean & \textbf{84.1} & \textbf{93.6} & \textbf{94.0} & 3.83  \\
    & Ft Cls-Landmarks-Clean $\Rightarrow$  Ft Rnk-Landmarks-Clean & 83.3 & 91.3 & 93.3 & 3.79  \\
\bottomrule
\end{tabular}
\label{tab:ml}
\end{table*}

\begin{figure*}[th!]
\centering
 \includegraphics[width=1\linewidth]{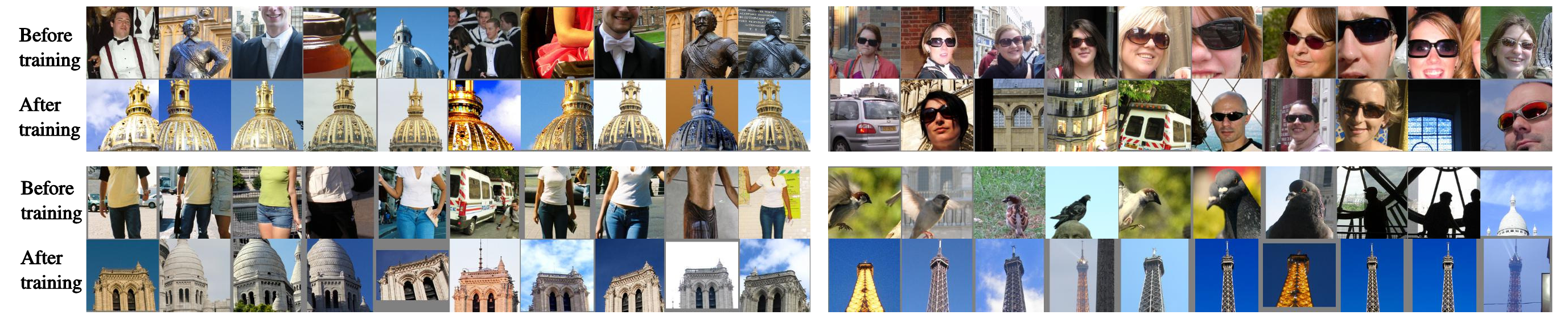}
 \caption{\label{fig:activations} Visualization of the neuron adaptation during training. Image patches with largest activation values for some neurons of layer ``conv5\_3'' from VGG16, before (top) and after (bottom) finetuning for retrieval. See text for more details.}
 \end{figure*}

\paragraph{Impact of finetuning on the neurons.}
We qualitatively evaluate the impact of finetuning the representation for retrieval. To this end, we visualize the image patches that most strongly respond (i.e. with the largest activation values) for different neurons of the last convolutional VGG16 layer in Fig.~\ref{fig:activations}, before and after finetuning for retrieval. These examples illustrate the process that takes place during finetuning. Some neurons that were originally specialized in recognizing specific object parts crucial for classification on ImageNet (for instance a ``shoulder neuron'' or a ``waist neuron'') were repurposed to fire on visually similar landmark parts (\eg domes, buildings with flat roofs and two windows). However, other neurons (\eg the ``sunglasses neuron'') were not clearly repurposed, which suggest that improvements in the training scheme may be possible.
 
\paragraph{Computational cost.}
To train and test our models we use an M40 NVIDIA GPU with 12 Gb of memory.
When pretraining ResNet101 on Landmarks-full with a classification loss it takes approximately 4 days to perform 80,000 iterations with a batch size of 128. This is the model that we use to initialize our approach in most of the experiments.
For training our ranking model we use a batch size of 64 triplets and the ``single stream'' approach (Algorithm \ref{alg:sst}), and resize our images preserving their aspect ratio such that the longer size has 800 pixels. With ResNet101 this process requires about 7.5 Gb of memory per stream per sample, and can process 64 iterations in approximately one hour, of which approximately 15 minutes are devoted to mining hard triplets. 
Our model, when initialized with the pretrained classification model, converges after approximately 3,000 iterations, \ie, 2 days. If we do not warm up the model on Landmarks and use directly the ImageNet model, it converges after approximately 8,000 iterations. In both cases, this roughly corresponds to a week of total training.

Once trained, extracting the descriptor of one image takes approximately 150 ms, \ie, about 7 images per second on a single GPU. Computing the similarity between two images comes down to computing the dot-product between their representations, which is very efficient, \ie, one can compute millions of such comparisons per second on a standard processor.

 %

\section{Improving the R-MAC representation}
\label{sec:part2}

The R-MAC representation has proved to excel at retrieval among deep methods \citep{Tolias2016}. In the previous section we have shown that we could further improve its effectiveness by learning the network weights in an end-to-end manner with an objective and a training set tailored for image retrieval. 
In this section, we propose several ways to modify the network architecture itself.
First, we improve the region pooling mechanism by introducing a region proposal network (RPN) that predicts the most relevant regions of the image, where the local features should be extracted (Section \ref{sec:proposal}). Second, we observe that the network architecture only considers a single fixed image resolution, and propose to extend it to build a multi-resolution descriptor (Section \ref{sec:multi-res}).

\subsection{Beyond fixed regions: proposal pooling}
\label{sec:proposal}
The rigid multi-scale grid used in R-MAC to pool regions tries to ensure that the object of interest is covered by at least one of the regions.
However, this raises two issues. First, it is unlikely that any of the grid regions precisely align with the object of interest.
Second, many of the regions only cover background, especially if the object to retrieve is of small scale. This is problematic as the comparison between R-MAC signatures can be seen as a many-to-many region matching, and so region clutter will negatively affect the performance. Increasing the number of regions in the grid would improve the likelihood that one region is well aligned with the object of interest, but would also increase the number of irrelevant regions.

\begin{figure*}[th!]
\includegraphics[width=1\linewidth]{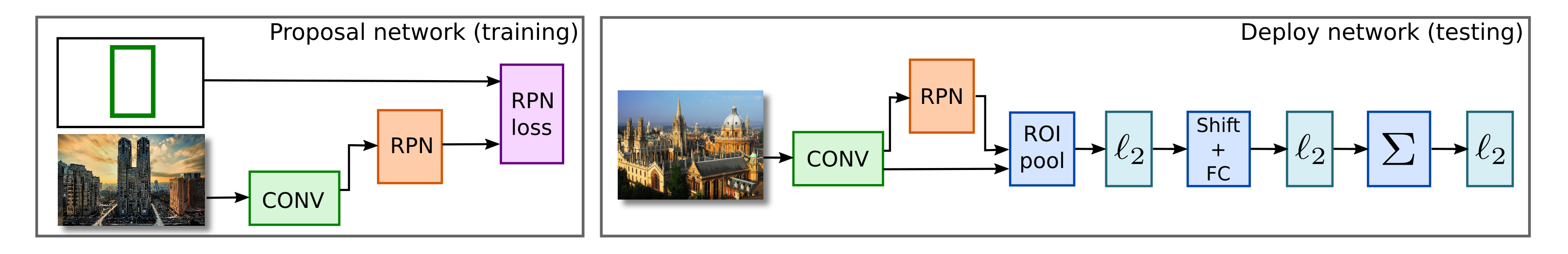}
\caption{\textbf{Proposal network.} 
At train time, a region proposal network is trained using bounding box annotations and an appropriate loss (left). At test time, the query image is fed to the learned architecture to
efficiently produce a \textit{compact global image representation} that can be compared with the dataset image representations with a
simple dot-product (right). \label{fig:rpn}}
\end{figure*}

We propose to modify the R-MAC architecture to enhance it with the ability to focus on relevant regions in the image. To this end we replace the rigid grid with a region proposal network (RPN) trained to localize regions of interest in images, similar to the proposal mechanism of \cite{Ren2015faster}. This RPN is trained using the approximate bounding box annotations of the Landmarks dataset obtained as a by-product of our cleaning process.
The resulting network architecture is illustrated in Fig.~\ref{fig:rpn}. 

The main idea behind an RPN is to predict, for a set of candidate boxes of various sizes and aspect ratios, a score describing how likely each candidate box at each possible image location contains an object of interest.
 Simultaneously, for each candidate box, it performs coordinate regression to improve the location accuracy.
This is achieved by a ``fully-convolutional'' network consisting of a first layer that uses $3\times 3$ filters, and two sibling convolutional layers with $1\times 1$ filters that predict, for each candidate box in the image and for each location, both the objectness score and the regressed coordinates. Non-maximum suppression is then performed on the ranked boxes to produce $k$ final proposals per image that are used to replace the rigid grid.

This modification to the network has several positive outcomes. First, the region proposals typically cover the object of interest more tightly than the rigid grid. Second, even if they do not overlap exactly with the region of interest, most of the proposals do overlap significantly with it, which means that increasing the number of proposals per image not only helps to increase the coverage but also helps in the many-to-many matching.

\paragraph{Learning the RPN.} We assign a binary class label to each candidate box, depending on how much
this box overlaps with the ground truth region of interest, and we minimize an objective function with a multitask loss that combines a
classification loss (more precisely a log loss over object \emph{vs} background classes)
and a regression loss (similar to the smooth $\ell_1$ loss used by \cite{Girshick2015}). The objective function is optimized by backpropagation and SGD.
More details about the implementation and the training procedure of the RPNs can be found in the work of \cite{Ren2015faster}.

We learn the RPN on top of the convolutional layers of our network. We first train the network using the rigid grid as described in Section \ref{sec:learning}, and then we fix the weights of the convolutional layers and train the RPN from the output of the last convolutional layer. In this way, both networks share the computation of the convolutional part and are combined into a single architecture (Fig.~\ref{fig:rpn}). Finetuning the RPN \emph{and} the ranking simultaneously is also feasible, but we observed no accuracy gain by doing so. 


\subsection{Multi-resolution}
\label{sec:multi-res}


In the original R-MAC descriptor as proposed by \cite{Tolias2016}, images are considered at a single scale. However, one could consider extracting and combining features from images that have been resized to different resolutions, in order to integrate information from different scales. The goal is to improve the matching between objects that appear at different scales in the database images and the retrieval of small objects.

One interesting characteristic of the original R-MAC network is that different input image sizes still produce descriptors of the same length. Note, however, that two versions of the same image with different resolutions will \emph{not} produce the same output descriptor. The first part of the network is fully convolutional, which directly enables to process inputs of different sizes, and the aggregation layer combines the size-dependent amount of input features into a fixed-length representation. Following this idea, we propose to extract different descriptors from images resized at different scales, and then combine them into a single final representation. 
In practice we use 3 scales, with 550, 800 and 1,050 pixels in the larger side, preserving the aspect ratio. The descriptor of each of the three images is then extracted independently. Finally we sum-aggregate them and $\ell_2$-normalize them to obtain the final descriptor.

This multi-resolution descriptor can be computed both in the query side and in the database side.
The process brings an extra computational cost at feature extraction time (approximately three times the cost for three resolutions), but the cost at search time and the storage cost remain the same.

Our multi-resolution scheme can be connected to previous papers aiming to build transformation-invariant representations like \cite{a,b}. The transformation considered in our case is the image scaling. In contrast to multi-column networks or bagging approaches \citep{a}, we use the same network for all image scales. In fact, our approach is conceptually close to \cite{b}, a siamese network with weight sharing, the main difference being that we use average-pooling instead of max-pooling.


\subsection{Experiments}

In this section we study the impact of the proposal pooling and the multi-resolution descriptors.

\paragraph{Experimental details.}
We train the RPN network for $200$k iterations with a weight decay of $5\cdot 10^{-5}$ and a learning rate of $10^{-3}$, which is decreased by a factor of $10$ after $100$k iterations.
We remark that only the RPN layers are updated and that the preceding convolutional layers remain fixed. 
The process takes less than $12$ hours on an M40 GPU.

\paragraph{Region proposal network.}

Table \ref{tab:proposals} presents the results of the region proposal network for an increasing number of regions compared to a rigid grid both for the baseline R-MAC (convolution weights learned from ImageNet) and for the version trained with a ranking loss, for both VGG16 and ResNet101 architectures.
With VGG16 we observe a significant improvement for all datasets and types of training when the number of regions is high enough (128 regions or more), consistent with our findings in the preliminary version of this article \citep{gordo2016deep}. 
However, with ResNet101, this gap is much smaller, especially when the network has been trained with the ranking loss.
Our intuition is that ResNet101 is able to learn a more invariant representation of the regions and to discount the effect of background, and so it does not require the proposals as much as VGG16. This makes the use of proposals less appealing when using ResNet101.
Given that ResNet101 considerably outperforms VGG16 for all the cases (\cf Tables \ref{tab:ml} and \ref{tab:proposals}), we depart from \cite{gordo2016deep} and, for the rest of the paper, we report results only with ResNet101 without using the RPN.

\begin{table*}[t!]
 \footnotesize
 \caption{Accuracy comparison between the fixed-grid and our proposal network, for an increasingly large number of proposals, before and after finetuning with a ranking-loss. The rigid grid extracts, on average, 20 regions per image.}
 \centering
 \begin{tabularx}{\textwidth}{@{}p{0.25cm}p{1.5cm}p{4.5cm}p{1.4cm}YYYYYY@{}}
 \toprule
 && & & \multicolumn{6}{c}{\bfseries \# Region Proposals} \\
 \cmidrule{5-10} 
 &Dataset & Model & {\bfseries Grid} & 20 &  32 & 64 & 128 & 192 & 256\\
 \midrule
 \rtext{VGG16}{9} &\multirow{2}{*}{Oxford 5k} & ILSVRC2012 baseline & 60.3 & 62.4 & 63.1 & 63.3 & 64.3 & 65.0 & \textbf{65.4} \\
 &   & Ft Cls-Full $\Rightarrow$  Ft Rnk-Clean & 79.9 & 80.7 & 80.8 & 81.9 & 83.1 & \textbf{83.2} & \textbf{83.2}\\
 \cmidrule{2-10}
 & \multirow{2}{*}{Paris 6k} & ILSVRC2012 baseline & 79.9 & 77.6 & 78.5 & 79.7 & 80.6 & 81.1 & \textbf{81.3}\\
 &    & Ft Cls-Full $\Rightarrow$  Ft Rnk-Clean  & 85.9 & 85.1 & 85.7 & 86.6 & 87.1 & 87.1 & \textbf{87.2}\\
 \cmidrule{2-10}
 & \multirow{2}{*}{Holidays} & ILSVRC2012 baseline & 85.8 & 82.7 & 83.5 & 85.8 & 86.8 & \textbf{87.5} & \textbf{87.5}\\
 &   & Ft Cls-Full $\Rightarrow$  Ft Rnk-Clean  & 87.9 & 86.2 & 86.7 & 87.8 & \textbf{88.7} & \textbf{88.7} & \textbf{88.7} \\
 \cmidrule{2-10}
 & \multirow{2}{*}{UKB} & ILSVRC2012 baseline & 3.75 & 3.72 & 3.74 & 3.76 & 3.77 & \textbf{3.78} & \textbf{3.78} \\ 
 &   & Ft Cls-Full $\Rightarrow$  Ft Rnk-Clean & 3.59 & 3.55 & 3.58 & 3.60 & 3.61 & \textbf{3.62} & \textbf{3.62} \\
 \midrule
  \rtext{ResNet101}{9}& \multirow{2}{*}{Oxford 5k} & ILSVRC2012 baseline & 69.4 & 69.2 & 70.5 & 71.4 & 72.3 & 72.5 & \textbf{72.9} \\
 &   & Ft Cls-Full $\Rightarrow$  Ft Rnk-Clean & 84.1 &   83.7 & 84.1 & 84.4 & 85.0 & \textbf{85.2} & \textbf{85.2}\\ 
 \cmidrule{2-10}
 & \multirow{2}{*}{Paris 6k} & ILSVRC2012 baseline & 85.2 & 84.5 & 85.2 & 86.0 & 86.3 & 86.5 & \textbf{86.6} \\
 &  & Ft Cls-Full $\Rightarrow$  Ft Rnk-Clean & 93.6 & 93.3 & 93.7 & \textbf{94.0} & \textbf{94.0} & \textbf{94.0} & \textbf{94.0} \\
 \cmidrule{2-10}
 & \multirow{2}{*}{Holidays} & ILSVRC2012 baseline & 91.4 & 89.8 & 91.0 & 92.0 & 92.2 & \textbf{92.3} & 92.2\\
 &  & Ft Cls-Full $\Rightarrow$  Ft Rnk-Clean & 94.0 & 92.0 & 92.7 & 93.5 & 93.8 & \textbf{94.0} & \textbf{94.0} \\
 \cmidrule{2-10}
 &  \multirow{2}{*}{UKB} & ILSVRC2012 baseline & 3.89 & 3.89 & 3.89 & 3.90 & 3.90 & \textbf{3.90} & \textbf{3.90} \\
 &    & Ft Cls-Full $\Rightarrow$  Ft Rnk-Clean & 3.83 & 3.82 & 3.82 & 3.83 & 3.83 & \textbf{3.84} & 3.83  \\
 \bottomrule
 \end{tabularx}
 \label{tab:proposals}
\end{table*}

\paragraph{Multi-resolution.}
\label{sub:multires}
Table \ref{tab:mr} shows results using ResNet101 trained with a ranking loss. Multi-resolution is applied to the query image (QMR), to the database images (DMR), or to both of them. All cases improve over the single-resolution descriptors, showing that encoding images using several scales helps at matching and retrieving objects.
QMR and DMR also appear to be complementary. We use both QMR and DMR through the rest of our experiments.

\begin{table}[t!]
 \footnotesize
 \caption{\textbf{Multi-resolution.} Effect of using multi-resolution descriptors on the query side (QMR) and on the database side (DMR).}
 \centering
 \begin{tabular}{cccccc}
\hline
QMR & DMR & Oxford 5k & Paris 6k & Holidays & UKB\\
\midrule
 &  & 84.1 & 93.6 & 94.0 & 3.83  \\
\cmark& & 84.9 & 94.1 & 94.3 & 3.83 \\
 &\cmark& 85.2 & 94.1 & 94.4 & 3.83  \\
\cmark&\cmark& \textbf{86.1} & \textbf{94.5} & \textbf{94.8} & \textbf{3.84}  \\
 \bottomrule
 \end{tabular}
 \label{tab:mr}
\end{table}

\section{Evaluation of the complete approach}
\label{sec:sota}

In the previous sections we have cast the R-MAC descriptor as a standalone network architecture where its weights can be learned discriminatively in an end-to-end manner as well as proposed some improvements over the original pipeline.
In this section we compare the obtained representation with the state of the art. Our final method integrates two other improvements: query expansion (QE) and database-side feature augmentation (DBA).

\subsection{Query expansion}
\label{sub:qe}
To improve the retrieval results we use query expansion, a standard technique introduced to the image search problem by \cite{Chum2007}. 
Query expansion works as follows: a first query is issued with the representation of the query image, and the top $k$ results are retrieved.
Those top $k$ results may then undergo a spatial verification stage, where results that do not match the query are discarded.
The remaining results, together with the original query, are then sum-aggregated and renormalized.
Finally, a second query is issued with the combined descriptor, producing the final list of retrieved images.
Query expansion typically leads to large improvements in accuracy at the expense of two extra costs at query time: spatial verification, and a second querying operation.
In our case we do not perform spatial verification (note that this typically requires access to local keypoint descriptors, which we do not have), and therefore query expansion simply doubles the query time due to the second query operation.

\subsection{Database-side feature augmentation}
\label{sub:dba}
Introduced in the works of \cite{Turcot2009} and \cite{Arandjelovic2012three}, database-side augmentation (DBA) replaces every image signature in the database by a combination of itself and its neighbors, potentially after a spatial verification stage as in the case of query expansion. The objective is to improve the quality of the image representations by leveraging the features of their neighbors.
Since we do not use spatial verification, we sum-aggregate the nearest $k$ neighbors as in the query expansion case.
Optionally, the sum can be weighted depending on the rank of the neighbors, and in our experiments we use $\text{weight}(r) = \frac{k - r}{k}$ as a weighting scheme, with $r$ the rank of the neighbor, and $k$ the total number of considered neighbors.

DBA is less common than query expansion as, with sparse inverted files, it increases the size of the database as well as the query time. In our case, signatures are already dense, so we are not affected by this. Consequently, the only extra cost incurs in finding the nearest neighbors in the dataset, which is done only once, and offline.
In the case of growing databases, the database augmentation could potentially be also done online as new samples are added.

\subsection{Experiments}
\label{sub:part3xps}

\subsubsection{Evaluation of QE and DBA}
\begin{figure}[t!]
 \begin{centering}
\includegraphics[height=4.3cm,trim={0.28cm 0 0.3cm 0},clip]{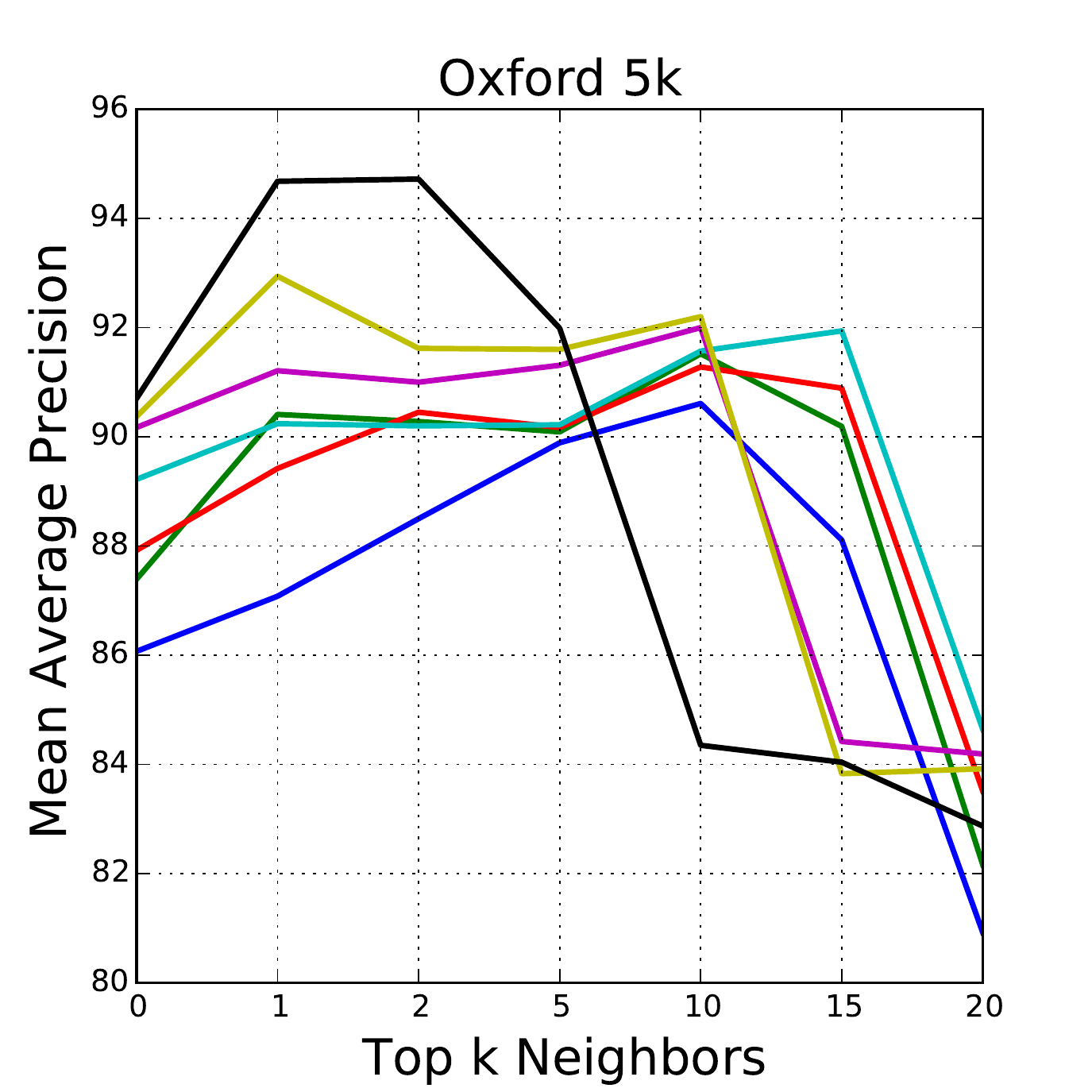}
\includegraphics[height=4.3cm,trim={0.65cm 0 0.3cm 0},clip]{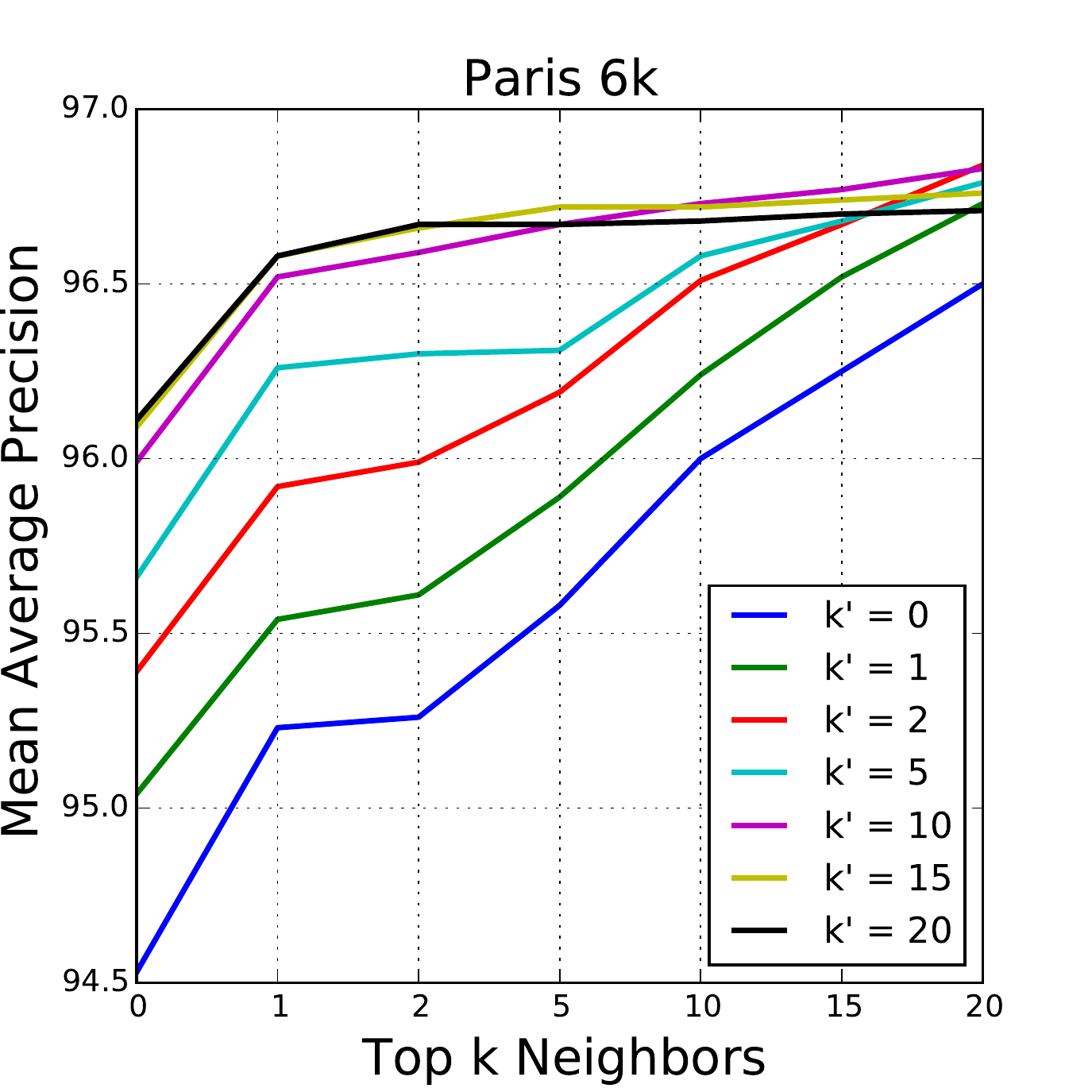}
 \par\end{centering}
 \caption{\label{fig:aqe} Accuracy as a function of the number of neighbors $k$ used during query expansion (QE) for several values of the number of neighbors $k'$ used for database-side augmentation (DBA).}
 \end{figure}

We evaluate the effect of query expansion (QE) depending on the number of neighbors $k$ as well as the effect of database-side augmentation (DBA) depending on the number of neighbors $k'$ in Fig. \ref{fig:aqe}.
First of all we observe how, in Oxford 5k, where many queries have very few relevant items (less than 10 or even less than 5), using large values of $k$ for the QE can, unsurprisingly, degrade the accuracy instead of improving it, independently of whether DBA is used or not. This is not a problem on Paris, where all queries have a large number of relevant items in the dataset. 

The weighted DBA seems to help in all cases, even when large values of $k'$ are selected, but, as a side effect, it can worsen the results as well if an inappropriate number of neighbors are chosen for QE.
In general it seems that QE and DBA can significantly help each other if the appropriate number of neighbors is chosen, and, as a rule of thumb, we suggest to use a large value for DBA (\eg $k'=20$) and a small value for QE (\eg $k=1$ or $k=2$). 
Because DBA can be a costly preprocessing, it is not always feasible. In this case (corresponding to $k'=0$), it is preferable to use an intermediate value for $k$.
For our final experiments involving QE and DBA, we fix $k=1$ and $k'=20$ in all datasets. When employing only QE we fix $k=10$ in all datasets. If one has prior knowledge about the dataset, modifying these values may lead to improved results.

\subsubsection{Comparison with the state of the art}
\begin{table*}[ht!]
   \renewcommand{\arraystretch}{0.8} 
 \caption{Accuracy comparison with the  state of the art. Methods marked with an * use the full image as a query in Oxford and Paris instead of using the annotated region of interest as is standard practice. The $^\dagger$ symbol denotes our reimplementation. Methods that manually rotate the images on Holidays using an oracle are labeled with $\triangleright$. We do not perform QE on Holidays as it is not a standard practice. See text for more details.\label{tab:soaresnet}}
 \footnotesize
 \centering
 \begin{tabularx}{\textwidth}{@{}p{0.25cm}p{4.5cm}p{0.8cm}YYYYY@{}} \toprule
 & &  & \multicolumn{5}{c}{\bfseries Datasets} \\
 \cmidrule{4-8} 
 & \bfseries{Method} & {\bfseries Dim.}  & Oxf5k & Par6k & Oxf105k & Par106k & Holidays \\
 \midrule 
 \rtext{Global descriptors}{18}&{~\cite{jegou:2014}}   &{  1024} & 56.0  &-    &  50.2    & -  & 72.0\\
 &~\cite{jegou:2014}   & { 128}  & 43.3  & -    &  35.3    & -        & 61.7  \\
 &~\cite{Gordo2012}  & { 512}  & -  & -  &  -   & -        & 79.0 \\
 &~\cite{Babenko2014} & { 128}  & 55.7*  & -    &  52.3*  & -        & 75.9/78.9$^{\triangleright}$  \\
 &~\cite{Gong2014}   & { 2048}  & -  & -  &  -   & - & 80.8 \\
 &~\cite{Babenko2015}   & { 256}  & 53.1  & -    &  50.1    & -        & 80.2$^{\triangleright}$  \\
 &~\cite{Ng2015}   & { 128}  & 59.3*  & 59.0*  &  -   & -        & 83.6 \\
 &~\cite{Paulin2015}   & { 256K}  & 56.5  & -  &  -   & -        & 79.3 \\
 &~\cite{Perronnin2015}   & { 4000}  & -  & -  &  -   & -        & 84.7 \\
 &~\cite{Tolias2016}  & { 512}  &66.9&83.0&61.6 &75.7 & 85.2$^{\dagger}$/86.9$^{\dagger,\triangleright}$\\
 &~\cite{Tolias2016} (ResNet101)$^\dagger$ & 2048 & 69.4 & 85.2 & 63.7 & 77.8 & 91.3 $^{\triangleright}$\\ 
 &~\cite{Kalantidis2016} & { 512}& 68.2  & 79.7 & 63.3  & 71.0 &  84.9 \\
 &~\cite{Arandjelovic2016} & { 4096}& 71.6 & 79.7 & - & - & 83.1/87.5$^{\triangleright}$  \\
 &~\cite{Radenovic2016} & { 512}& 79.7 & 83.8 & 73.9 & 76.4 & 82.5$^{\triangleright}$ \\
 \cmidrule{2-8} 
 & Previous state of the art & & 79.7 \cite{Radenovic2016} & 83.8 \cite{Radenovic2016} & 73.9 \cite{Radenovic2016} & 76.4 \cite{Radenovic2016} & 84.9 \cite{Kalantidis2016}\\
 \cmidrule{2-8} 
 & {\footnotesize \textbf{Ours}} & { 2048}  & \textbf{86.1} & \textbf{94.5} & \textbf{82.8} & \textbf{90.6} & 90.3/\textbf{94.8$^{\triangleright}$}\\ 
 \midrule
 \midrule
 \rtext{Matching / Spatial verif. / QE}{18}& ~\cite{Chum2011} &&       82.7   & 80.5  &     76.7     &    71.0 & -  \\
 &~\cite{Danfeng2011} &&    81.4   &     80.3  &76.7    &    - 	 &  -    \\
 &~\cite{Mikulik2013} && 84.9 &    82.4  & 79.5 &    77.3   & 75.8$^{\triangleright}$   \\
 &~\cite{Shen2014} &&        75.2  &      74.1  &     72.9     &     -     & 76.2 \\
 &~\cite{Tao2014} &&        77.8  &       -    &       -      &       -      & 78.7 \\
 &~\cite{Deng2013}   &&  84.3  & 83.4  &  80.2   & -        & 84.7 \\
 &~\cite{Tolias2015}  & &  86.9  &     85.1  &85.3     &   - &   81.3	 \\
 &~\cite{Tolias2016} & { 512} & 77.3  &86.5 &  73.2    &79.8 & -  \\
 &~\cite{Tolias2016} (ResNet101)$^\dagger$ & 2048 & 78.9 & 89.7 & 75.5 &  85.3 &   \\
 &~\cite{Tolias2015b} &&89.4 & 82.8 & 84.0  & - & - \\
 &~\cite{Xinchao2015} &&73.7  & - & -  & - & 89.2 \\
 &~\cite{Kalantidis2016} & { 512}& 72.2  & 85.5 & 67.8  & 79.7 &  - \\
 &~\cite{Radenovic2016} & { 512}& 85.0 &86.5 & 81.8 & 78.8  & - \\
 &~\cite{Azizpour2015} &  & 79.0 & 85.1 & -  & -  & \textbf{90.0} \\
 \cmidrule{2-8} 
 & Previous state of the art & & 89.4 \cite{Tolias2015b} & 86.5 \cite{Tolias2016} & 85.3 \cite{Tolias2015} & 79.8 \cite{Tolias2016} & \textbf{90.0} \cite{Azizpour2015}\\
 \cmidrule{2-8} 
 &{\footnotesize \textbf{Ours (with QE)}} & { 2048} & 90.6 & 96.0 & 89.4 & 93.2 & - \\ 
 &{\footnotesize \textbf{Ours (with QE and DBA)}} & { 2048} & \textbf{94.7} & \textbf{96.6} & \textbf{93.6} & \textbf{93.5} & -\\ 
 \bottomrule
 \end{tabularx}
 \end{table*}

We compare our method against the state of the art in Table \ref{tab:soaresnet}.
For these experiments, in addition to the four datasets introduced in Section \ref{sec:exp-siamese}, we also consider the \textbf{Oxford 105k} and \textbf{Paris 106k}
datasets that extend Oxford 5k and Paris 6k with 100k distractor images \citep{Philbin2007}.
In the first half of the table, we show results for other methods that employ global representations of images and do not perform any form of spatial verification or query expansion at run-time.
As such, they are conceptually closer to our method.
Yet, we consistently outperform all of them on all datasets. 
In one case (namely, on Paris 106k), our method is more than 14 mAP points ahead of the best competitor \citep{Radenovic2016}.

The de facto evaluation protocol for methods based on CNN features on the Holidays dataset involves manually rotating the images to correct their orientation.
If we do not manually rotate the images, our accuracy drops from 94.8 to 90.3, which still outperforms the current state of the art.
Instead of using an oracle to rotate the database images, one can automatically rotate the query image and issue three different queries (original query, query rotated 90 degrees, and query rotated 270 degrees). The score of one database image is the maximum score obtained with the three queries. This makes the query process 3 times slower, but improves the accuracy to 92.9 with no oracle intervention. 

We also include our reimplementation of the R-MAC baseline \citep{Tolias2016} using ResNet101 instead of VGG16.
Although the accuracy improvement when using ResNet101 is not negligible, the accuracy obtained by the trained model is still much higher (in Oxford, 69.4 without training vs 84.1 and 86.1 when training, either using single-resolution or multi-resolution testing). 
This gap underlines the importance of both a well designed architecture and a sound end-to-end training with relevant data, all tailored to the particular task of image retrieval.

The second part of Table \ref{tab:soaresnet} shows results for state-of-the-art methods that do not necessarily rely on a global representation. 
The majority of them is characterized by a larger memory footprint than our method, \eg the ones of \cite{Tolias2015b, Tolias2016, Danfeng2011, Azizpour2015}. These methods perform a costly spatial verification at runtime that typically requires storing thousands of local descriptors for each image in the database \citep{Tolias2015b, Xinchao2015, Mikulik2013}. 
Most of them also perform query expansion (QE). For comparison purposes, we also report our results using QE with or without  DBA at the bottom of the table. Using only QE brings about half of the improvement obtained when using both QE and DBA, yet avoiding any pre-processing of the database. In spite of not requiring any form of spatial verification at runtime, our method is
 largely improving on the state of the art on all datasets. In particular, our performance is between 5 to 14 mAP points ahead of the best competitor on all datasets.
 
The best methods in the literature \citep{Tolias2015b, Azizpour2015} are hardly scalable as they require a lot of storage memory and an expensive verification. For instance, the method of \cite{Tolias2015b} requires a slow spatial verification taking over 1 second per query (excluding descriptor extraction time). Without spatial verification their approach loses 5 mAP points and still requires about 200 ms per query. The approach of \cite{Tolias2016} is more scalable but still needs an extra spatial-verification stage based on storing many local representations of the database images, ending up in a significantly larger memory footprint than our approach, despite using advanced compression techniques. 
In comparison, our approach only calculates two matrix-vector products (only one if QE is not performed), that are extremely efficient.
This operation computes several millions of image comparisons in less than a second. Without any compression, our method requires storing 2,048 floats per image, \ie 8 kb, but this representation can be drastically compressed without much accuracy loss as we show in the next section.
Finally, we would like to point out that, when not performing QE and DBA (that leverage information about the target dataset at test time), our method uses a single universal model -- the same for all test datasets -- contrary to, for instance, other methods of \cite{Danfeng2011, Shen2014, Tolias2015} that perform some learning on the target datasets.

We also report results on the UKB dataset using our universal model. Our method obtains 3.84 recall@4 without QE and DBA, and 3.91 recall@4 score with QE and DBA. The latter is comparable to the best published results on this dataset, \ie 3.85 reported by \cite{Azizpour2015}, although this method is a lot more costly.
Other results are significantly lower (\eg \cite{Paulin2015} reports 3.76, \cite{Deng2013} reports 3.75, and \cite{Tolias2015b} reports 3.67) and they are hardly scalable as well (see discussion above). 
Note that training marginally decreases our performance on UKB (Table \ref{tab:ml}). This is caused by the discrepancy between our training set (landmarks images) and the UKB images (daily life items). The drop remains marginal, which suggests that out method adapts well to other retrieval contexts.

\subsubsection{Short image codes with PCA and PQ}
\label{sub:shortcodes}

\begin{figure*}[ht!]
  \begin{centering}
    \includegraphics[height=3.98cm,trim={0.28cm 0 1.1cm 0},clip]{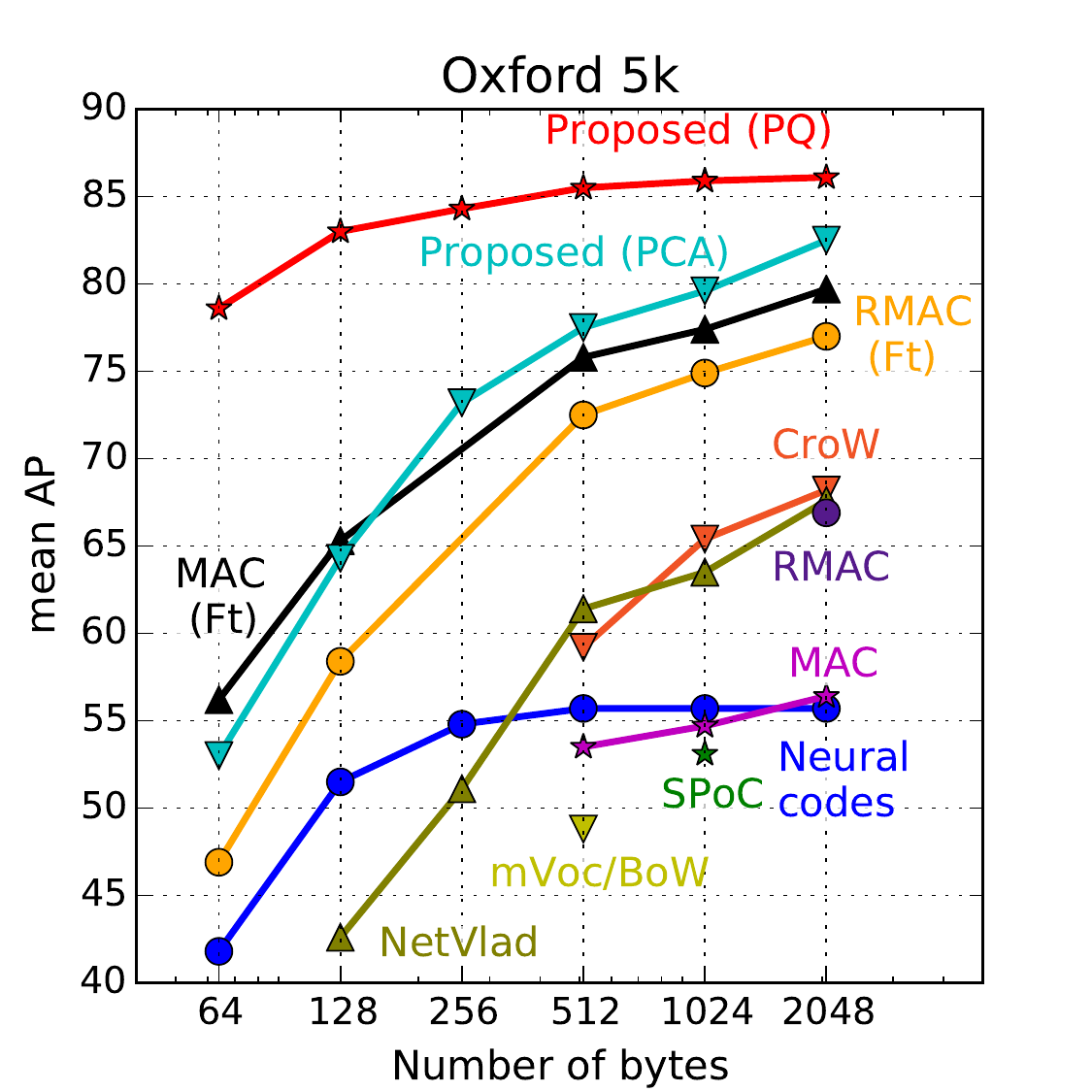}
    \includegraphics[height=3.98cm,trim={0.68cm 0 1.1cm 0},clip]{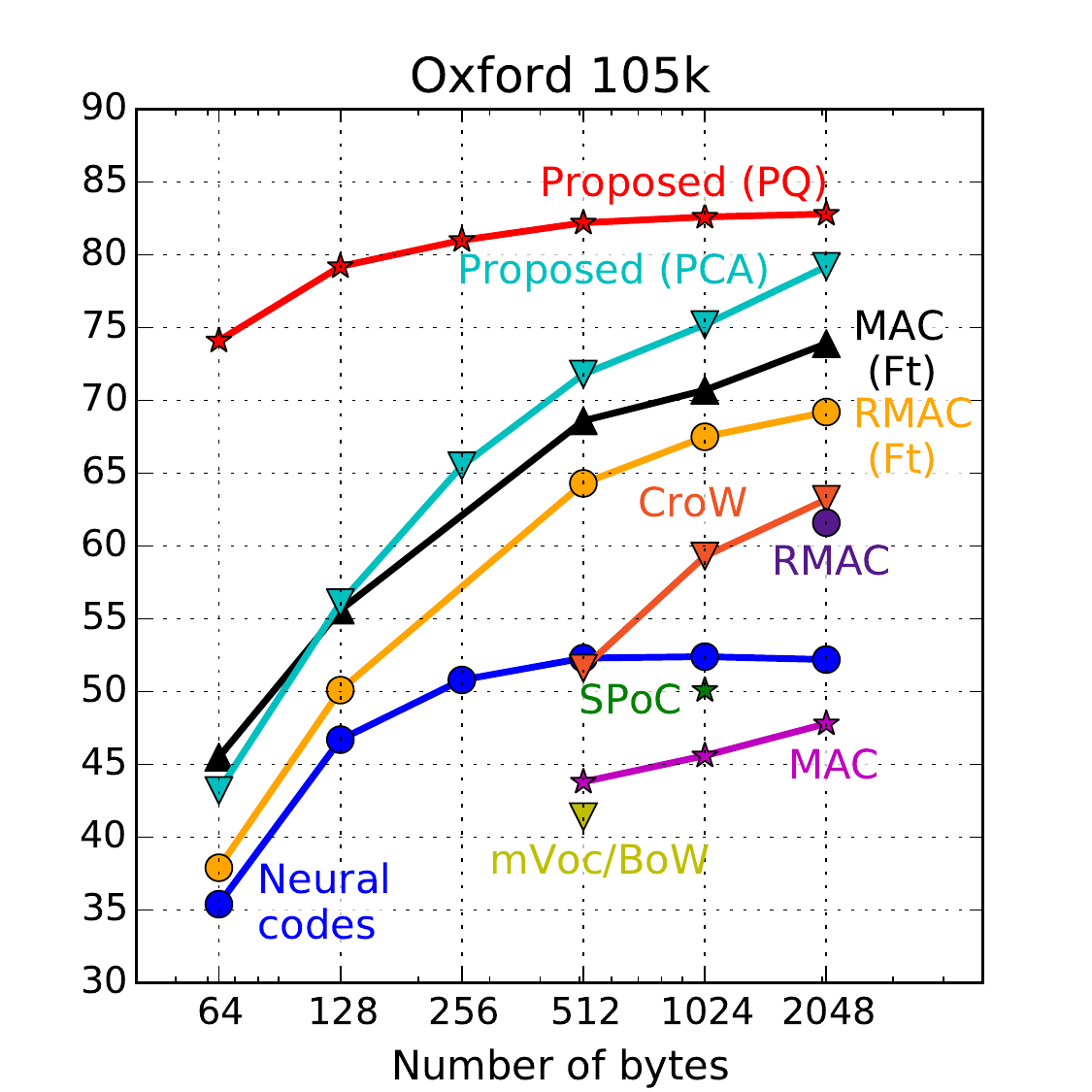}
    \includegraphics[height=3.98cm,trim={0.68cm 0 1.1cm 0},clip]{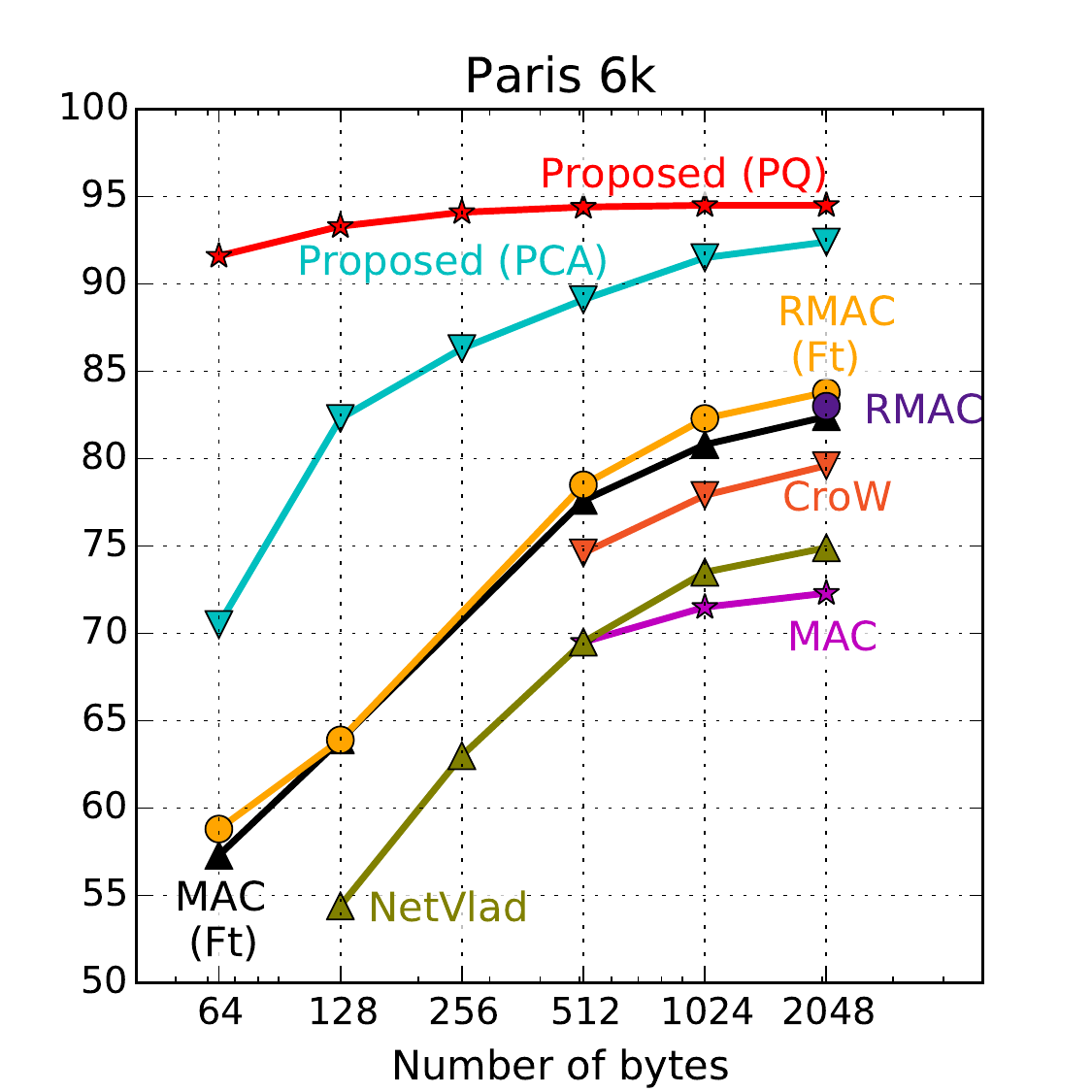}
    \includegraphics[height=3.98cm,trim={0.68cm 0 1.1cm 0},clip]{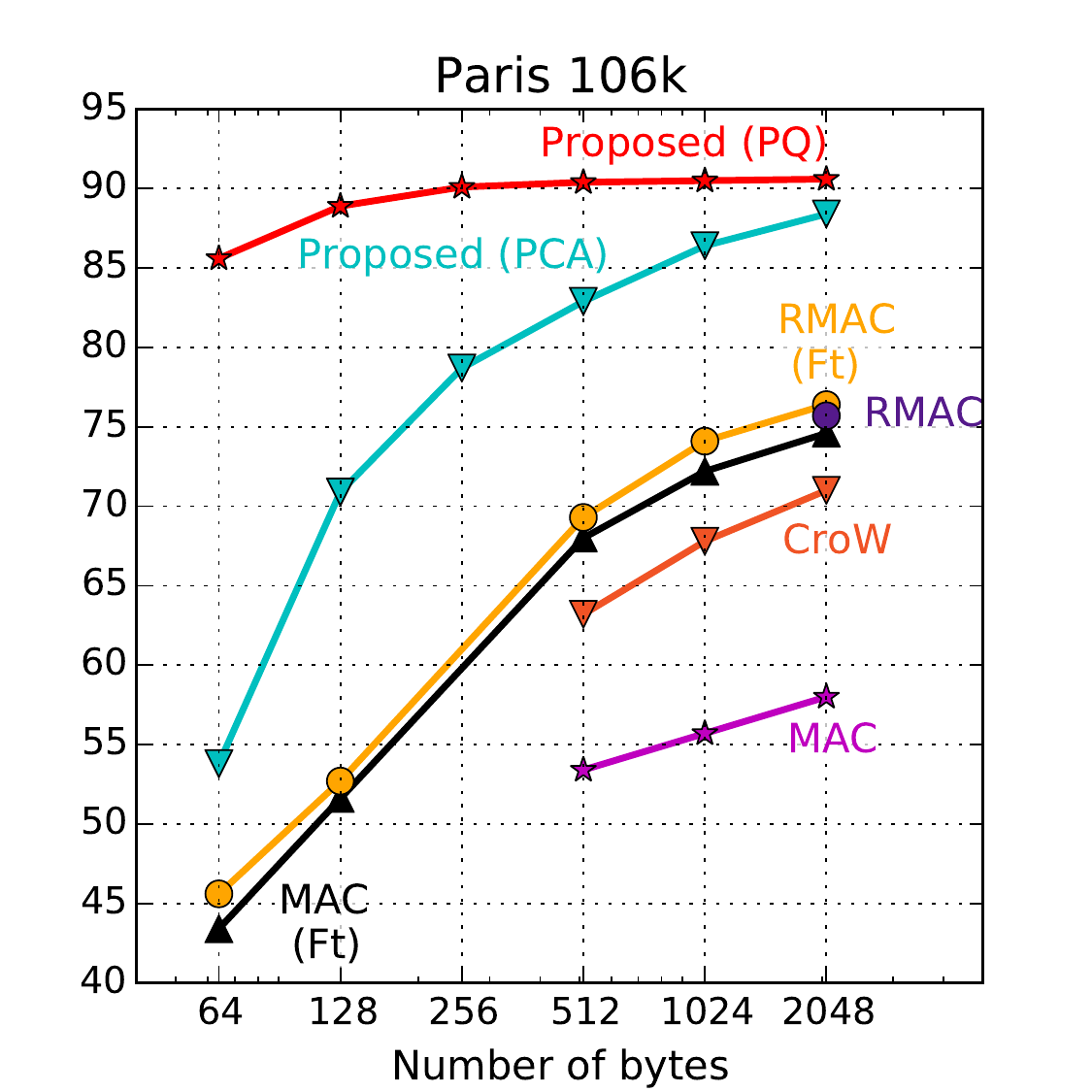}
    \includegraphics[height=3.98cm,trim={0.68cm 0 1.1cm 0},clip]{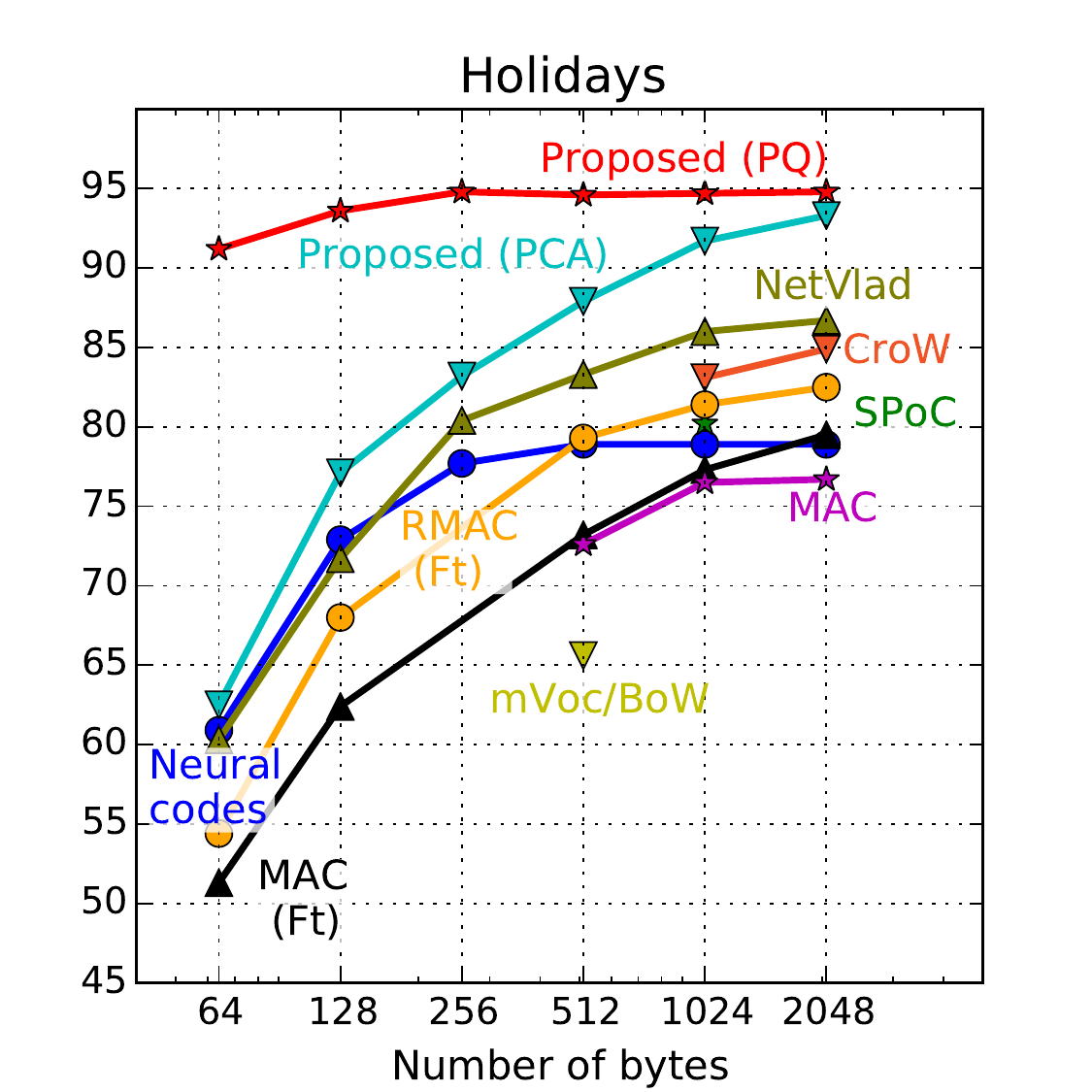}
 \par\end{centering}
 \caption{\label{fig:pq}Results for short image codes. Our method with PQ and PCA compression, compared to finetuned MAC and R-MAC \citep{Radenovic2016}, CroW \citep{Kalantidis2016}, MAC and R-MAC \citep{Tolias2016}, Neural codes \citep{Babenko2014}, NetVlad \citep{Arandjelovic2016}, SPoC \citep{Gong2014}, and mVOC/BoW \citep{Radenovic2015}.}
\end{figure*}

We investigate two different methods to reduce the memory footprint of our approach while preserving the best possible accuracy.
We compress our 2048-dimensional image descriptors 
using either principal component analysis (PCA) or product quantization (PQ) \citep{PQ2011}.
In both cases, we learn the vocabulary (PCA projection or PQ codebook) on Landmarks-clean images, encoded with our learned representation. 

In the case of PCA, to obtain descriptors of $d$ dimensions we simply mean center the features, project them with the eigenvectors associated with the $d$ largest eigenvalues of the data, and $\ell_2$-normalize them. 
The resulting descriptor size is thus $4d$ bytes, as they are stored as 32-bits floats.
PQ compression, for its part, is based on splitting the input descriptor in $k$ subparts and applying vector quantization on each subpart separately.
Although some works also apply PCA to the input descriptors before the PQ encoding, we found it did not have any noticeable impact in our case.
Training PQ is then equivalent to learning a codebook for each subpart and is achieved though k-means clustering on a set of representative descriptors. 
The codebook size is typically set to 256 for each subpart, as it allows them to be stored on exactly 1 byte. Thus, the size of a PQ-encoded descriptor is $k$ bytes.
At test time, efficient caching techniques allow computing the dot-product between the query and the PQ-encoded database descriptors efficiently \citep{PQ2011}. 
Note that recent improvements have led PQ to match the high speed of bitwise Hamming distance computations without losing in accuracy \citep{Douze2016}.

Retrieval results for our method (without QE or DBA) and for the state of the art are presented in Fig.~\ref{fig:pq} for all datasets and for different descriptor sizes (in bytes).
PCA-based compression, labeled as ``Proposed (PCA)'', achieves slightly better results than other existing approaches for all considered datasets and all code sizes, but its accuracy drops rapidly for short codes.
This compression method is still of interest as it does not require any change in the system architecture and still 
compares favorably to the state of the art. 
PQ-based compression, labeled as ``Proposed (PQ)'' in Fig~\ref{fig:pq}, largely outperforms all published methods in terms of the performance versus size trade-off by a large margin, on all datasets. 
Even for very short image codes of 64 bytes, it is able to outperform most of the state of the art that uses codes of 2,048 bytes.
In this setting, we can store hundreds of millions of images on a single machine with 64 Gb of RAM, which demonstrates the scalability of our approach.

\subsection{Qualitative results}\label{s:qualitative}
\begin{figure*}[ht!]
 \begin{centering}
     \includegraphics[width=1\linewidth]{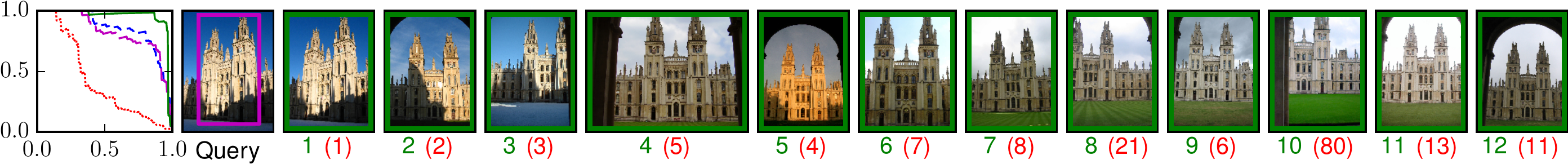}
     \includegraphics[width=1\linewidth]{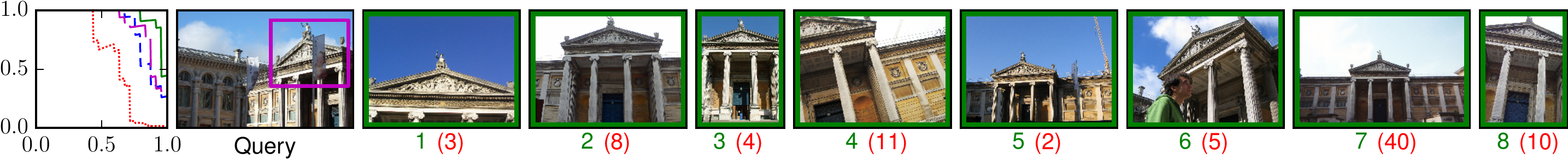}
     \includegraphics[width=1\linewidth]{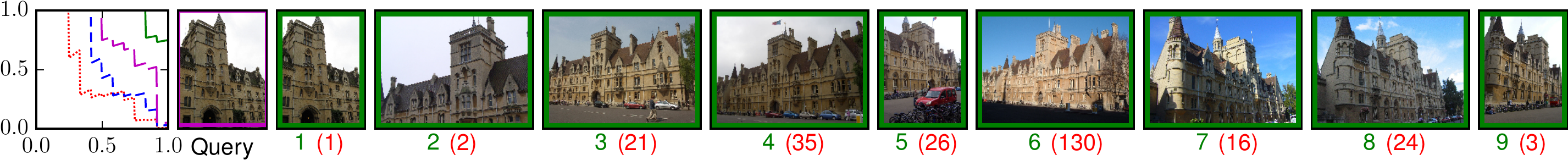}
     \includegraphics[width=1\linewidth]{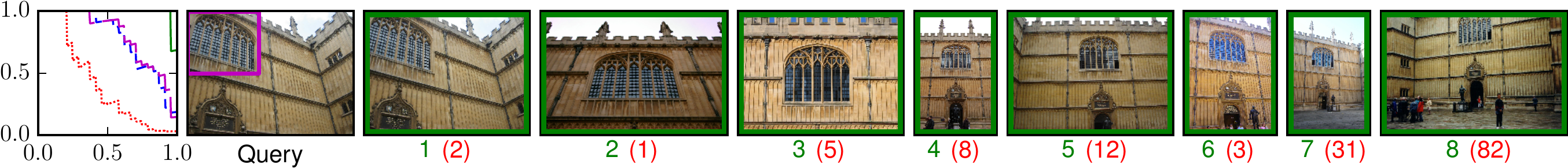}
     \includegraphics[width=1\linewidth]{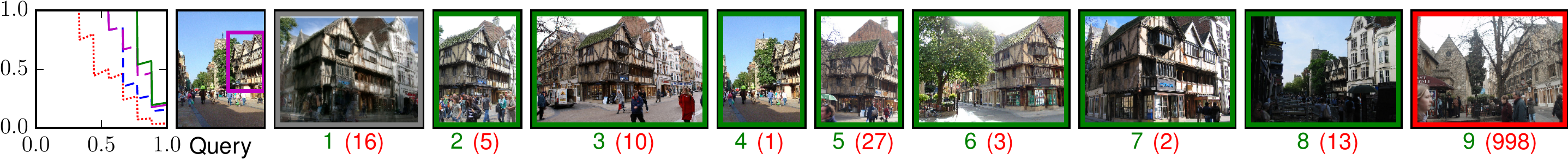}
     \includegraphics[width=1\linewidth]{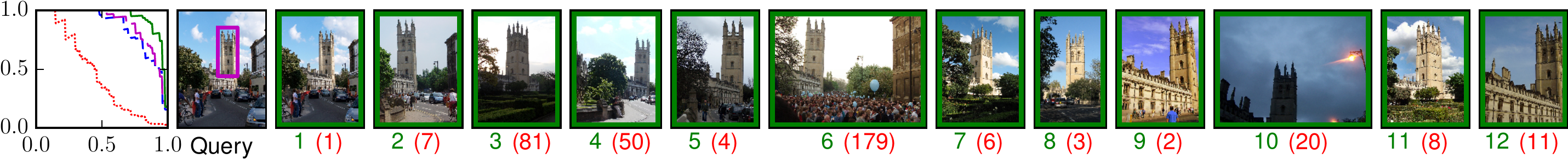}
     \includegraphics[width=1\linewidth]{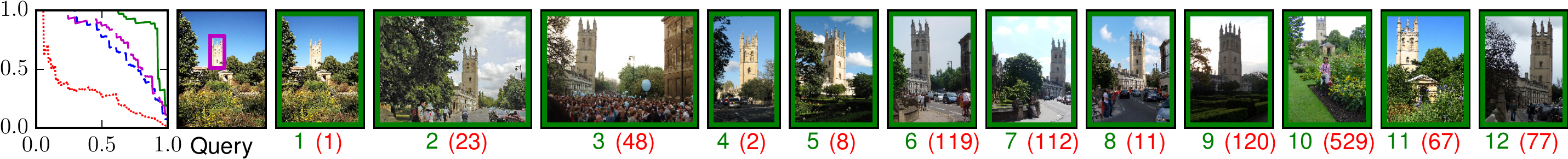}
 \par\end{centering}
 \caption{\label{fig:ox-qual} Top retrieved images and AP curves for a
   few Oxford queries (purple rectangle from leftmost images). On the
   plot, R-MAC baseline is red, our learned version is blue,
   multi-resolution is purple, and the full system with QE and DBA is
   green. Green, gray and red borders on images respectively denote
   positive, null and negative images.}
 \vspace{-0.3cm}
 \end{figure*}

Fig. \ref{fig:ox-qual} shows the top retrieved images by our final best performing retrieval system based on ResNet101 (including QE and DBA) on some Oxford 5k queries (purple rectangle on the leftmost images). For every query we also provide the corresponding average precision (AP) curve (green curve) and compare it with the ones obtained for the baseline R-MAC (red curve), our learned architecture (blue curve), and its multi-resolution flavor (purple curve). The results obtained with the proposed trained model are consistently better in terms of accuracy.
In many cases, several of the correctly retrieved images by our method were not well scored by the baseline method, that placed them far down in the list of results.

\section{Conclusions}
 \label{sec:conclusions}

 We have presented an effective and scalable method for instance-level image retrieval that encodes images into compact global signatures that can be compared with the dot-product.
 The proposed approach combines three ingredients that are key to success. First, we gathered a suitable training set by automatically cleaning an existing landmarks dataset. Second, we proposed a learning framework that relies on a triplet-based ranking loss, and that leverages this training set to train a deep architecture. Third, for the deep architecture, we built on the R-MAC descriptor, cast it as a fully differentiable network so we could learn its weights, and enhanced it with a proposal network that focuses on the most relevant image regions.
 Extensive experiments on several benchmarks show that our representation significantly outperforms the state of the art when using global signatures, even when using short codes of $64$ or $128$ bytes.
Our method also outperforms the state of the art set by more complex methods that rely on costly matching and verification, and does so while being faster and more memory-efficient.
 



\bibliographystyle{spbasic}      
\bibliography{egbib}   

\begin{thebibliography}{74}
\providecommand{\natexlab}[1]{#1}
\providecommand{\url}[1]{{#1}}
\providecommand{\urlprefix}{URL }
\expandafter\ifx\csname urlstyle\endcsname\relax
  \providecommand{\doi}[1]{DOI~\discretionary{}{}{}#1}\else
  \providecommand{\doi}{DOI~\discretionary{}{}{}\begingroup
  \urlstyle{rm}\Url}\fi
\providecommand{\eprint}[2][]{\url{#2}}

\bibitem[{Antol et~al(2015)Antol, Agrawal, Lu, Mitchell, Batra, Zitnick, and
  Parikh}]{Antol2015VQA}
Antol S, Agrawal A, Lu J, Mitchell M, Batra D, Zitnick CL, Parikh D (2015) Vqa:
  Visual question answering. In: ICCV

\bibitem[{Arandjelovic and Zisserman(2012)}]{Arandjelovic2012three}
Arandjelovic R, Zisserman A (2012) Three things everyone should know to improve
  object retrieval. In: CVPR

\bibitem[{Arandjelovic et~al(2016)Arandjelovic, Gronat, Torii, Pajdla, and
  Sivic}]{Arandjelovic2016}
Arandjelovic R, Gronat P, Torii A, Pajdla T, Sivic J (2016) Net{VLAD}: {CNN}
  architecture for weakly supervised place recognition. In: CVPR

\bibitem[{Azizpour et~al(2015)Azizpour, Razavian, Sullivan, Maki, and
  Carlsson}]{Azizpour2015}
Azizpour H, Razavian A, Sullivan J, Maki A, Carlsson S (2015) Factors of
  transferability for a generic convnet representation. IEEE Transactions on
  Pattern Analysis and Machine Intelligence PP(99):1--1

\bibitem[{Babenko and Lempitsky(2015)}]{Babenko2015}
Babenko A, Lempitsky VS (2015) Aggregating deep convolutional features for
  image retrieval. In: ICCV

\bibitem[{Babenko et~al(2014)Babenko, Slesarev, Chigorin, and
  Lempitsky}]{Babenko2014}
Babenko A, Slesarev A, Chigorin A, Lempitsky VS (2014) Neural codes for image
  retrieval. In: ECCV

\bibitem[{Chopra et~al(2005)Chopra, Hadsell, and Lecun}]{Chopra2005}
Chopra S, Hadsell R, Lecun Y (2005) Learning a similarity metric
  discriminatively, with application to face verification. In: In Proc. of
  Computer Vision and Pattern Recognition Conference

\bibitem[{Chum et~al(2007)Chum, Philbin, Sivic, Isard, and
  Zisserman}]{Chum2007}
Chum O, Philbin J, Sivic J, Isard M, Zisserman A (2007) Total recall: Automatic
  query expansion with a generative feature model for object retrieval. In:
  ICCV

\bibitem[{Chum et~al(2011)Chum, Mikulik, Perdoch, and Matas}]{Chum2011}
Chum O, Mikulik A, Perdoch M, Matas J (2011) Total recall {II}: Query expansion
  revisited. In: CVPR

\bibitem[{Danfeng et~al(2011)Danfeng, Gammeter, Bossard, Quack, and
  Van~Gool}]{Danfeng2011}
Danfeng Q, Gammeter S, Bossard L, Quack T, Van~Gool L (2011) Hello neighbor:
  accurate object retrieval with k-reciprocal nearest neighbors. In: CVPR

\bibitem[{Deng et~al(2013)Deng, Ji, Liu, Tao, and Gao}]{Deng2013}
Deng C, Ji R, Liu W, Tao D, Gao X (2013) Visual reranking through weakly
  supervised multi-graph learning. In: ICCV

\bibitem[{Deng et~al(2009)Deng, Dong, Socher, Li, Li, and Fei-Fei}]{Deng2009}
Deng J, Dong W, Socher R, Li LJ, Li K, Fei-Fei L (2009) {ImageNet: A
  Large-Scale Hierarchical Image Database}. In: CVPR

\bibitem[{Douze et~al(2016)Douze, Jegou, and Perronnin}]{Douze2016}
Douze M, Jegou H, Perronnin F (2016) Polysemous codes. In: ECCV

\bibitem[{Frome et~al(2013)Frome, Corrado, Shlens, Bengio, Dean, Ranzato, and
  Mikolov}]{frome13devise}
Frome A, Corrado GS, Shlens J, Bengio S, Dean J, Ranzato MA, Mikolov T (2013)
  Devise: A deep visual-semantic embedding model. In: NIPS

\bibitem[{Girshick(2015)}]{Girshick2015}
Girshick R (2015) Fast {R-CNN}. In: CVPR

\bibitem[{Girshick et~al(2014)Girshick, Donahue, Darrell, and
  Malik}]{Girshick2014}
Girshick R, Donahue J, Darrell T, Malik J (2014) Rich feature hierarchies for
  accurate object detection and semantic segmentation. In: CVPR

\bibitem[{Gong et~al(2014)Gong, Wang, Guo, and Lazebnik}]{Gong2014}
Gong Y, Wang L, Guo R, Lazebnik S (2014) Multi-scale orderless pooling of deep
  convolutional activation features. In: ECCV

\bibitem[{Gordo et~al(2012)Gordo, Rodr{\'{\i}}guez{-}Serrano, Perronnin, and
  Valveny}]{Gordo2012}
Gordo A, Rodr{\'{\i}}guez{-}Serrano JA, Perronnin F, Valveny E (2012)
  Leveraging category-level labels for instance-level image retrieval. In: CVPR

\bibitem[{Gordo et~al(2016)Gordo, Almaz{\'{a}}n, Revaud, and
  Larlus}]{gordo2016deep}
Gordo A, Almaz{\'{a}}n J, Revaud J, Larlus D (2016) Deep image retrieval:
  Learning global representations for image search. In: ECCV

\bibitem[{Hadsell et~al(2006)Hadsell, Chopra, and Lecun}]{Hadsell2006}
Hadsell R, Chopra S, Lecun Y (2006) Dimensionality reduction by learning an
  invariant mapping. In: CVPR

\bibitem[{Hays and Efros(2008)}]{Hays:08}
Hays J, Efros AA (2008) im2gps: estimating geographic information from a single
  image. In: CVPR

\bibitem[{He et~al(2014)He, Zhang, Ren, and Sun}]{He2014}
He K, Zhang X, Ren S, Sun J (2014) Spatial pyramid pooling in deep
  convolutional networks for visual recognition. In: ECCV

\bibitem[{He et~al(2016)He, Zhang, Ren, and Sun}]{He2016}
He K, Zhang X, Ren S, Sun J (2016) Deep residual learning for image
  recognition. In: CVPR

\bibitem[{Hoffer and Ailon(2015)}]{Hoffer2015}
Hoffer E, Ailon N (2015) Deep metric learning using triplet network. In: SIMBAD

\bibitem[{Hu et~al(2014)Hu, Lu, and Tan}]{Hu2014}
Hu J, Lu J, Tan YP (2014) Discriminative deep metric learning for face
  verification in the wild. In: CVPR

\bibitem[{J{\'e}gou and Chum(2012)}]{Jegou2012}
J{\'e}gou H, Chum O (2012) Negative evidences and co-occurences in image
  retrieval: The benefit of pca and whitening. In: ECCV

\bibitem[{J\'egou and Zisserman(2014)}]{jegou:2014}
J\'egou H, Zisserman A (2014) Triangulation embedding and democratic
  aggregation for image search. In: CVPR

\bibitem[{J\'egou et~al(2008)J\'egou, Douze, and Schmid}]{Jegou2008}
J\'egou H, Douze M, Schmid C (2008) Hamming embedding and weak geometric
  consistency for large scale image search. In: ECCV

\bibitem[{J{\'e}gou et~al(2010)J{\'e}gou, Douze, and Schmid}]{Jegou2010}
J{\'e}gou H, Douze M, Schmid C (2010) Improving bag-of-features for large scale
  image search. IJCV

\bibitem[{J\'egou et~al(2010)J\'egou, Douze, Schmid, and
  P\'erez}]{Jegou2010aggregating}
J\'egou H, Douze M, Schmid C, P\'erez P (2010) Aggregating local descriptors
  into a compact image representation. In: CVPR

\bibitem[{Jegou et~al(2011)Jegou, Douze, and Schmid}]{PQ2011}
Jegou H, Douze M, Schmid C (2011) Product quantization for nearest neighbor
  search. TPAMI

\bibitem[{Kalantidis et~al(2016)Kalantidis, Mellina, and
  Osindero}]{Kalantidis2016}
Kalantidis Y, Mellina C, Osindero S (2016) Cross-dimensional weighting for
  aggregated deep convolutional features. In: Workshop on Web-scale Vision and
  Social Media (VSM), ECCV

\bibitem[{Karpathy et~al(2014)Karpathy, Joulin, and Fei-Fei}]{karpathy14deep}
Karpathy A, Joulin A, Fei-Fei L (2014) Deep fragment embeddings for
  bidirectional image-sentence mapping. In: NIPS

\bibitem[{Krizhevsky et~al(2012)Krizhevsky, Sutskever, and
  Hinton}]{Krizhevsky2012}
Krizhevsky A, Sutskever I, Hinton G (2012) Image{N}et classification with deep
  convolutional neural networks. In: NIPS

\bibitem[{Laptev et~al(2016)Laptev, Savinov, Buhmann, and Pollefeys}]{b}
Laptev D, Savinov N, Buhmann JM, Pollefeys M (2016) Ti-pooling:
  Transformation-invariant pooling for feature learning in convolutional neural
  networks. In: CVPR

\bibitem[{Li et~al(2015)Li, Larson, and Hanjalic}]{Xinchao2015}
Li X, Larson M, Hanjalic A (2015) Pairwise geometric matching for large-scale
  object retrieval. In: CVPR

\bibitem[{Long et~al(2015)Long, Shelhamer, and Darrell}]{Long2015}
Long J, Shelhamer E, Darrell T (2015) Fully convolutional networks for semantic
  segmentation. In: CVPR

\bibitem[{Lowe(2004)}]{Lowe2004}
Lowe DG (2004) Distinctive image features from scale-invariant keypoints. IJCV

\bibitem[{Makadia et~al(2008)Makadia, Pavlovic, and Kumar}]{Makadia:08}
Makadia A, Pavlovic V, Kumar S (2008) A new baseline for image annotation. In:
  ECCV

\bibitem[{Mikolajczyk and Schmid(2004)}]{Mikolajczyk2004}
Mikolajczyk K, Schmid C (2004) Scale \& affine invariant interest point
  detectors. IJCV

\bibitem[{Mikul{\'\i}k et~al(2010)Mikul{\'\i}k, Perdoch, Chum, and
  Matas}]{Mikulik2010}
Mikul{\'\i}k A, Perdoch M, Chum O, Matas J (2010) Learning a fine vocabulary.
  In: ECCV

\bibitem[{Mikulik et~al(2013)Mikulik, Perdoch, Chum, and Matas}]{Mikulik2013}
Mikulik A, Perdoch M, Chum O, Matas J (2013) Learning vocabularies over a fine
  quantization. IJCV

\bibitem[{Ng et~al(2015)Ng, Yang, and Davis}]{Ng2015}
Ng JYH, Yang F, Davis LS (2015) Exploiting local features from deep networks
  for image retrieval. In: CVPR workshops

\bibitem[{Nister and Stewenius(2006)}]{Nister2006}
Nister D, Stewenius H (2006) Scalable recognition with a vocabulary tree. In:
  CVPR

\bibitem[{Paulin et~al(2015)Paulin, Douze, Harchaoui, Mairal, Perronin, and
  Schmid}]{Paulin2015}
Paulin M, Douze M, Harchaoui Z, Mairal J, Perronin F, Schmid C (2015) Local
  convolutional features with unsupervised training for image retrieval. In:
  ICCV

\bibitem[{Perdoch et~al(2009)Perdoch, Chum, and Matas}]{Perdoch2009}
Perdoch M, Chum O, Matas J (2009) Efficient representation of local geometry
  for large scale object retrieval. In: CVPR

\bibitem[{Perronnin and Dance(2007)}]{Perronnin2007}
Perronnin F, Dance C (2007) Fisher kernels on visual vocabularies for image
  categorization. In: CVPR

\bibitem[{Perronnin and Larlus(2015)}]{Perronnin2015}
Perronnin F, Larlus D (2015) Fisher vectors meet neural networks: A hybrid
  classification architecture. In: CVPR

\bibitem[{Perronnin et~al(2010)Perronnin, Liu, S{\'a}nchez, and
  Poirier}]{Perronnin2010}
Perronnin F, Liu Y, S{\'a}nchez J, Poirier H (2010) Large-scale image retrieval
  with compressed fisher vectors. In: CVPR

\bibitem[{Philbin et~al(2007)Philbin, Chum, Isard, Sivic, and
  Zisserman}]{Philbin2007}
Philbin J, Chum O, Isard M, Sivic J, Zisserman A (2007) Object retrieval with
  large vocabularies and fast spatial matching. In: CVPR

\bibitem[{Philbin et~al(2008)Philbin, Chum, Isard, Sivic, and
  Zisserman}]{Philbin2008}
Philbin J, Chum O, Isard M, Sivic J, Zisserman A (2008) Lost in quantization:
  Improving particular object retrieval in large scale image databases. In:
  CVPR

\bibitem[{Philbin et~al(2010)Philbin, Isard, Sivic, and
  Zisserman}]{Philbin2010}
Philbin J, Isard M, Sivic J, Zisserman A (2010) Descriptor learning for
  efficient retrieval. In: ECCV

\bibitem[{Radenovic et~al(2015)Radenovic, Jegou, and Chum}]{Radenovic2015}
Radenovic F, Jegou H, Chum O (2015) Multiple measurements and joint
  dimensionality reduction for large scale image search with short
  vectors-extended version. In: International Conference on Multimedia
  Retrieval

\bibitem[{Radenovic et~al(2016)Radenovic, Tolias, and Chum}]{Radenovic2016}
Radenovic F, Tolias G, Chum O (2016) {CNN} image retrieval learns from {B}o{W}:
  Unsupervised fine-tuning with hard examples. In: ECCV

\bibitem[{Razavian et~al(2014)Razavian, Azizpour, Sullivan, and
  Carlsson}]{Razavian2014}
Razavian AS, Azizpour H, Sullivan J, Carlsson S (2014) {CNN} features
  off-the-shelf: an astounding baseline for recognition. In: CVPR Deep Vision
  Workshop

\bibitem[{Ren et~al(2015)Ren, He, Girshick, and Sun}]{Ren2015faster}
Ren S, He K, Girshick R, Sun J (2015) Faster {R-CNN}: Towards real-time object
  detection with region proposal networks. In: NIPS

\bibitem[{Rodriguez-Serrano et~al(2015)Rodriguez-Serrano, Larlus, and
  Dai}]{Rodriguez2015}
Rodriguez-Serrano J, Larlus D, Dai Z (2015) Data-driven detection of prominent
  objects. TPAMI

\bibitem[{Russakovsky et~al(2015)Russakovsky, Deng, Su, Krause, Satheesh, Ma,
  Huang, Karpathy, Khosla, Bernstein, Berg, and Fei-Fei}]{ILSVRC15}
Russakovsky O, Deng J, Su H, Krause J, Satheesh S, Ma S, Huang Z, Karpathy A,
  Khosla A, Bernstein M, Berg AC, Fei-Fei L (2015) {ImageNet Large Scale Visual
  Recognition Challenge}. IJCV

\bibitem[{Schmidhuber(2012)}]{a}
Schmidhuber J (2012) Multi-column deep neural networks for image
  classification. In: CVPR

\bibitem[{Schroff et~al(2015)Schroff, Kalenichenko, and Philbin}]{SchroffK2015}
Schroff F, Kalenichenko D, Philbin J (2015) Facenet: A unified embedding for
  face recognition and clustering. In: CVPR

\bibitem[{Shen et~al(2014)Shen, Lin, Brandt, and Wu}]{Shen2014}
Shen X, Lin Z, Brandt J, Wu Y (2014) Spatially-constrained similarity
  measurefor large-scale object retrieval. TPAMI

\bibitem[{Simo-Serra et~al(2015)Simo-Serra, Trulls, Ferraz, Kokkinos, Fua, and
  Moreno-Noguer}]{Serra2015}
Simo-Serra E, Trulls E, Ferraz L, Kokkinos I, Fua P, Moreno-Noguer F (2015)
  Discriminative learning of deep convolutional feature point descriptors. In:
  ICCV

\bibitem[{Simonyan and Zisserman(2015)}]{Simonyan2015verydeep}
Simonyan K, Zisserman A (2015) Very deep convolutional networks for large-scale
  image recognition. In: ICLR

\bibitem[{Sivic and Zisserman(2003)}]{Sivic2003}
Sivic J, Zisserman A (2003) Video google: A text retrieval approach to object
  matching in videos. In: ICCV

\bibitem[{Song et~al(2016)Song, Xiang, Jegelka, and Savarese}]{Song2015}
Song HO, Xiang Y, Jegelka S, Savarese S (2016) Deep metric learning via lifted
  structured feature embedding. In: CVPR

\bibitem[{Sun et~al(2014)Sun, Chen, Wang, and Tang}]{Sun2014}
Sun Y, Chen Y, Wang X, Tang X (2014) Deep learning face representation by joint
  identification-verification. In: NIPS

\bibitem[{Tao et~al(2014)Tao, Gavves, Snoek, and Smeulders}]{Tao2014}
Tao R, Gavves E, Snoek CG, Smeulders AW (2014) Locality in generic instance
  search from one example. In: CVPR

\bibitem[{Tolias and J\'egou(2015)}]{Tolias2015b}
Tolias G, J\'egou H (2015) Visual query expansion with or without geometry:
  refining local descriptors by feature aggregation. PR

\bibitem[{Tolias et~al(2015)Tolias, Avrithis, and J{\'e}gou}]{Tolias2015}
Tolias G, Avrithis Y, J{\'e}gou H (2015) Image search with selective match
  kernels: Aggregation across single and multiple images. IJCV

\bibitem[{Tolias et~al(2016)Tolias, Sicre, and J\'egou}]{Tolias2016}
Tolias G, Sicre R, J\'egou H (2016) Particular object retrieval with integral
  max-pooling of cnn activations. In: ICLR

\bibitem[{Torralba et~al(2008)Torralba, Fergus, and Freeman}]{tiny}
Torralba A, Fergus R, Freeman WT (2008) 80 million tiny images: a large dataset
  for non-parametric object and scene recognition. IEEE Trans PAMI

\bibitem[{Turcot and Lowe(2009)}]{Turcot2009}
Turcot P, Lowe DG (2009) Better matching with fewer features: The selection of
  useful features in large database recognition problems. In: ICCV Workshops

\bibitem[{Vardi and Zhang(2004)}]{GeoMedian2004}
Vardi Y, Zhang CH (2004) The multivariate {L1}-median and associated data
  depth. In: Proceedings of the National Academy of Sciences

\bibitem[{Wang et~al(2014)Wang, Song, Leung, Rosenberg, Wang, Philbin, Chen,
  and Wu}]{Wang2014}
Wang J, Song Y, Leung T, Rosenberg C, Wang J, Philbin J, Chen B, Wu Y (2014)
  Learning fine-grained image similarity with deep ranking. In: CVPR

\end{thebibliography}

\end{document}